# Optimal Scheduling Algorithms for LLM Inference: Theory and Practice

AGRIM BARI*, The University of Texas at Austin, United States
PARIKSHIT HEGDE*, The University of Texas at Austin, United States
GUSTAVO DE VECIANA, The University of Texas at Austin, United States

With the growing use of Large Language Model (LLM)-based tools like ChatGPT, Perplexity, and Gemini across industries, there is a rising need for efficient LLM inference systems. These systems handle requests with a unique two-phase computation structure: a prefill-phase that processes the full input prompt and a decode-phase that autoregressively generates tokens one at a time. This structure calls for new strategies for routing and scheduling requests.

In this paper, we take a comprehensive approach to this challenge by developing a theoretical framework that models routing and scheduling in LLM inference systems. We identify two key design principles—optimal tiling and dynamic resource allocation—that are essential for achieving high throughput. Guided by these principles, we propose the Resource-Aware Dynamic (RAD) scheduler and prove that it achieves throughput optimality under mild conditions. To address practical Service Level Objectives (SLOs) such as serving requests with different Time Between Token (TBT) constraints, we design the SLO-Aware LLM Inference (SLAI) scheduler. SLAI uses real-time measurements to prioritize decode requests that are close to missing their TBT deadlines and reorders prefill requests based on known prompt lengths to further reduce the Time To First Token (TTFT) delays.

We evaluate SLAI on the `openchat_shareGPT4` dataset using the Mistral-7B model on an NVIDIA RTX ADA 6000 GPU. Compared to Sarathi-Serve, SLAI reduces the median TTFT by 53% and increases the maximum serving capacity by 26% such that median TTFT is below 0.5 seconds, while meeting tail TBT latency constraints. The complete source code is available at: https://github.com/agrimUT/SLAI.

*Both authors contributed equally to this work.

Authors' Contact Information: Agrim Bari, agrimchaudhry@gmail.com, Chandra Department of Electrical and Computer Engineering, The University of Texas at Austin, Austin, TX, United States; Parikshit Hegde, parihegde96@gmail.com, Chandra Department of Electrical and Computer Engineering, The University of Texas at Austin, Austin, TX, United States; Gustavo de Veciana, deveciana@utexas.edu, Chandra Department of Electrical and Computer Engineering, The University of Texas at Austin, Austin, TX, United States.

## 1 Introduction

***LLM inference systems.*** The core problem in Large Language Model (LLM) inference is to generate a response autoregressively, one token[1] at a time, given a *prompt*—for example, "What is the capital of France?" producing the output "It is Paris." Modern LLMs such as GPT-4, Llama 3, and Gemini now power a wide range of services, including chatbots, coding assistants, and search engines. These services handle millions of user requests daily, and private deployments are rapidly increasing. As a result, there is growing interest in optimizing how requests are processed across one or more Graphics Processing Unit (GPU)-enabled nodes in data centers, since improved efficiency can lead to significant reductions in infrastructure and operating costs.

***Objectives for LLM serving systems.*** To meet growing demand, LLM systems must be carefully designed to make efficient use of hardware. A well-designed system keeps each active GPU busy—fully utilizing both its compute and memory—while also keeping response times low. This leads to two main goals: (1) achieving high throughput, measured in requests per second, to reduce the cost per request, and (2) maintaining low latency, which directly affects user experience. Latency is typically measured using two Service-Level Objectives (SLOs)[2]: *Time To First Token* (TTFT), which is the delay between a request's arrival and the generation of the first output token, capturing how long a user waits before the LLM starts responding; and *Time Between Tokens* (TBT), which is the time between successive output tokens, indicating the rate at which the response is streamed to the user. In real-world systems, request routing, scheduling, and caching are used to meet these goals. This paper focuses on *scheduling*, which plays a critical role in increasing throughput, reducing TTFT, and keeping TBT within acceptable limits.

***Phases of an LLM request and scheduler decisions.*** Each request to an LLM based on the Transformer architecture goes through two main phases: *prefill* and *decode*. In the *prefill-phase*, the model processes the entire prompt and generates the first output token. After that, the request enters the *decode-phase*, where it produces one token at a time in an auto-regressive manner until a `stop` token is generated. These two phases have distinct characteristics. The prefill-phase is highly parallelizable and can fully utilize the GPU's compute resources. By contrast, the decode phase is sequential and has low parallelism, which means that multiple decode-phase requests must be batched together to make efficient use of the GPU. Additionally, Transformer models store intermediate representations of tokens—called the *Key-Value (KV) cache*—which grow with the number of tokens processed and consume GPU memory. The scheduler's job is to select a mix of prefill-phase and decode-phase requests to include in each GPU batch. These decisions must balance GPU compute and memory bandwidth usage, stay within the memory budget, and meet latency SLOs.

***Challenges in scheduler design.*** Designing an effective scheduler for LLM inference presents several key challenges. First, GPU compute efficiency depends heavily on the composition of each batch—that is, the mix of prefill and decode-phase requests scheduled together. As a result, the scheduler must construct batches carefully to make the most of available resources. Second, the scheduler must balance resource allocation between the prefill and decode phases, as both phases have their respective SLOs: TTFT and TBT. Prioritizing prefill-phase requests can reduce TTFT but may delay decodes and worsen TBT. Conversely, prioritizing decode-phase requests keeps TBT low but can lower compute utilization and increase TTFT for new requests. Third, while the prompt length is known upon a request arrival the output length is unknown, making memory management complex—particularly since GPU memory is often a bottleneck. Finally, many LLM serving systems support multiple user tiers which have heterogeneous performance needs.

[1]A token in a LLM is a unit of text-such as a word, subword, or character-used as the basic input element for processing and generation.

[2]Thresholds may vary by application, but these two metrics are commonly used.

***Our approach.*** In this work, we approach the LLM scheduling problem from two complementary perspectives. From a theoretical standpoint, we develop a rigorous framework for analyzing and achieving throughput-optimal scheduling. From a practical perspective, we design a scheduler that dynamically adapts to diverse latency SLOs across heterogeneous user tiers.

***Contributions.*** Our key contributions are:

(1) *A Theoretical Model for LLM Inference Systems.* We develop a comprehensive analytical framework for request routing and scheduling in LLM inference systems, capable of capturing salient features of practical policies considered in the literature. Additionally, we introduce a model to represent inference computation times on modern GPU architectures. This unified framework enables rigorous analysis and comparison of routing and scheduling policies in such systems. See Section 3.
(2) *A Throughput Optimal Scheduler.* We identify two key design principles that characterize optimal resource-utilization in LLM inference systems; optimal tiling and optimal dynamic resource allocation across prefill and decode workloads. Guided by these insights, we design a simple load-balancing routing strategy and introduce a Resource-Aware Dynamic (RAD) scheduler. We rigorously prove that, under mild assumptions, this combination achieves throughput optimality. See Section 4.
(3) *Insights for Practical Systems.* Recognizing that in practical LLM inference systems it is desirable to meet latency SLOs, we provide insights into how real-world systems approximate and adapt the proposed design principles to satisfy these constraints. See Section 5.
(4) *A SLO-Aware Scheduler.* We design a practical scheduler called SLO-Aware LLM Inference (SLAI) scheduler that aims to minimize online median TTFT when serving requests with heterogeneous TBT constraints. To achieve this, SLAI uses online measurements to decide when the execution of a decode-iteration has become *time-critical*, i.e., should be prioritized for scheduling. In addition, it uses the known prompt length information to order prefill-phase requests so as to reduce the median TTFT. See Section 6.
(5) *Experimental Performance.* We evaluate SLAI on the `openchat_shareGPT4` dataset using the Mistral-7B LLM on an NVIDIA RTX ADA 6000 GPU. Our results show that SLAI can reduce the median TTFT by 53% while meeting TBT requirements compared to Sarathi-serve, the current state-of-the-art scheduler. Additionally, when median TTFT can not exceed 0.5 seconds, SLAI increases the serving capacity by 26%. See Section 7.

### 1.1 Related Work

LLM serving systems must make decisions about *when* to run the prefill and decode phases of a request, *where* to execute each request, and *how* to manage the growing KV cache produced by the model. Based on these challenges, prior work can be broadly categorized into three areas: (a) scheduling within a single inference node, (b) routing across multiple inference nodes, and (c) managing the KV cache. In this section, we focus exclusively on scheduling. For a discussion of related works on routing and KV cache management, we refer the reader to Appendix A.

*1.1.1 Queueing based analysis for schedulers.* The queueing-theoretic study of LLM inference remains in its early stages, especially when compared to the more mature systems-oriented literature. A recent survey by Mitzenmacher et al. [19] highlights the distinctive characteristics of LLM inference workloads and emphasizes the importance of queueing-based analysis for such systems. A closely related work is [17], in which the authors develop a queueing-theoretic model for LLM inference schedulers and analyze throughput optimality. The authors approximate the batch-processing time, which is observed to have a staircase-behaviour in practice, by a linear function. They observe under their model that work conserving schedulers are throughput optimal. In our work, we model the detailed computations on a GPU to perform LLM inference, and obtain a formula for the batch processing time that reflects the staircase-behaviour observed in practice. Our more detailed modeling framework yields novel structural insights—particularly on the optimal tiling and resource allocation strategies required for throughput optimality. These insights offer new avenues for improving the performance of practical LLM inference systems. In [20], the authors study throughput-maximization in a setting where the system is backed up with requests, and the token generation times depend on the batch size, but not on the token positions.

*1.1.2 Scheduling policies.* Schedulers differ in how they prioritize prefill-phase and decode-phase requests, and in their batching granularity. Below, we highlight some of the key work in this literature.

***Decode-prioritizing (request-level) schedulers.*** Frameworks such as FasterTransformer [23, 24] and the request-level mode of TensorRT-LLM process a set of requests till completion, before admitting any new requests. Since new prefill-phase requests never interrupt ongoing decode-phase requests, these schedulers perform well on TBT. However, they have low throughput when there is a imbalance in the total (prompt and output) length of requests and thus GPU may be under-utilized.

***Prefill-prioritizing (iteration-level) schedulers.*** Iteration-level batching, first introduced by Orca [33], enables dynamic admission and completion of requests at each forward pass. However, it relies on static memory allocation for the KV cache, which limits the number of concurrent requests to 16 on an A100 GPU. vLLM [15] overcomes this limitation using paged attention, allowing more flexible memory management and increasing the maximum number of concurrent requests to 128. Additionally, like FlashDecoding++ [8] and DeepSpeed-FastGen [7], vLLM aggressively admits new prefill-phase requests to improve throughput. However, this eager admission policy can delay decode iterations—especially for long prompts—leading to higher TBT latencies.

***Hybrid schedulers.*** Sarathi-Serve [2] introduces a token-budgeted, chunked-prefill strategy to balance throughput and TBT, effectively reducing decode-iteration stalls that are common in prefill-prioritizing schedulers. Beyond such scheduling-focused methods, recent systems propose orthogonal techniques aimed at improving overall LLM inference performance. BlendServe [34] targets offline workloads by reordering requests based on their resource usage profiles to improve hardware efficiency. POD-Attention [14] enables pipelined execution of prefill and decode phases to increase kernel overlap and improve GPU utilization. HydraGen [13] reduces redundant computation by identifying and merging shared prompt prefixes across requests. DistServe [36] adopts a disaggregated architecture that separates prefill and decode execution across different nodes, thereby eliminating intra-GPU contention; however, it introduces communication overhead due to the transfer of large KV caches.

## 2 Background

### 2.1 Transformer

We focus on the widely used *decoder-only* Transformer architecture—referred to simply as the Transformer in this work—that forms the basis of GPT, LLaMA, PaLM and other families of LLM models. The Transformer is an autoregressive model that generates token $\theta_{i+1}$ by conditioning on all previous tokens $\theta_1, \ldots, \theta_i$. Token $\theta_i$ is processed by the Transformer as shown in Fig. 1. A token is first converted into an embedding and then passed through $N$ layers. Let the input to the $n^{\text{th}}$ layer be a vector $\boldsymbol{x}^n \in \mathbb{R}^{d_x}$ of dimension $d_x$. Each layer comprises of three distinct sub-layers:

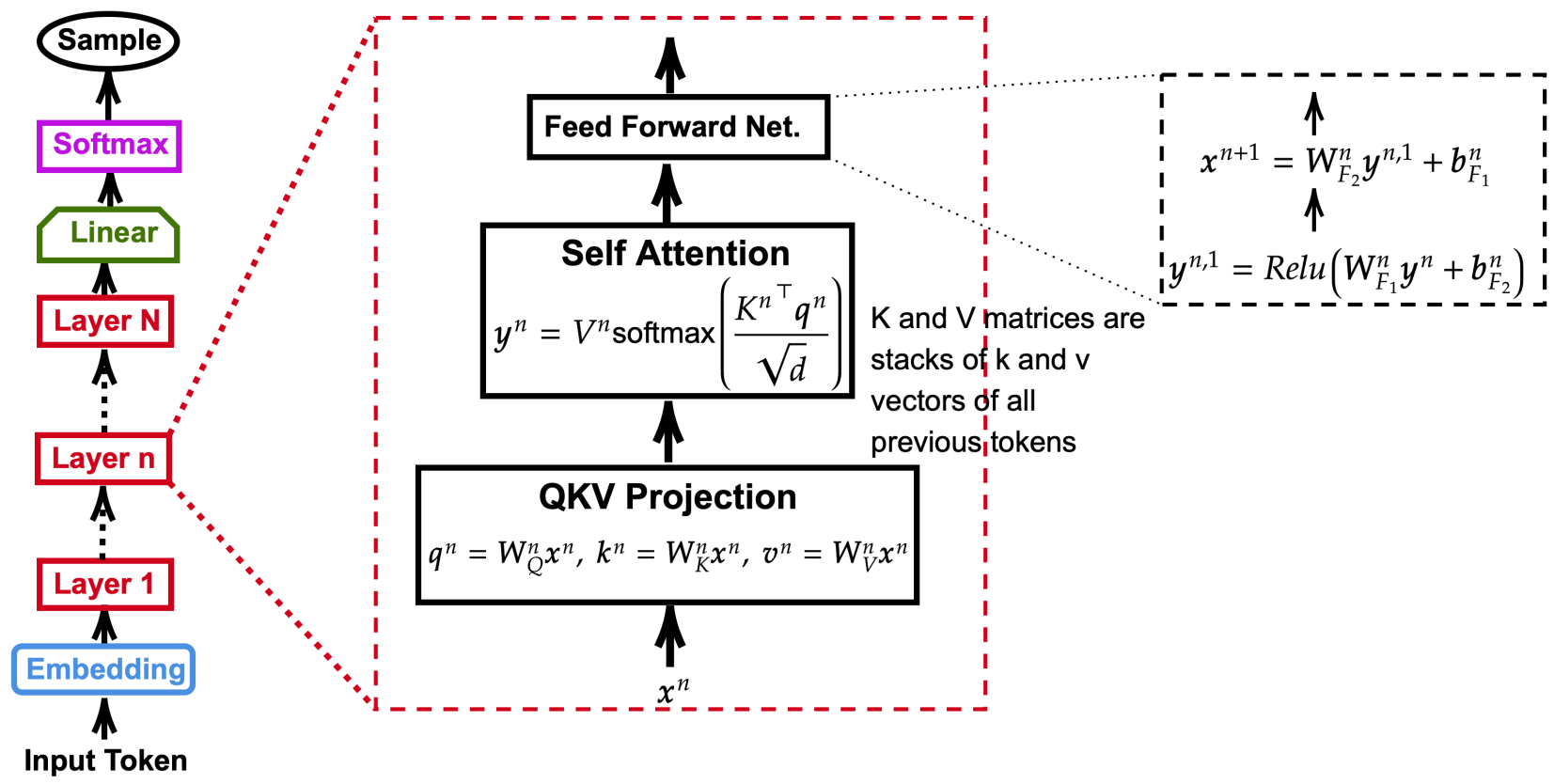

Fig. 1. Transformer architecture.

(1) *QKV Projection:* This sub-layer is parametrized by three matrices $W_Q^n$, $W_K^n$ and $W_V^n$ each of dimension $d \times d_x$ . The corresponding *Query*, *Key* and *Value* (or, QKV) vectors are computed using a linear transformation as, $\boldsymbol{q}^n = W_Q^n \boldsymbol{x}^n$, $\boldsymbol{k}^n = W_K^n \boldsymbol{x}^n$ and $\boldsymbol{v}^n = W_V^n \boldsymbol{x}^n$, where $\boldsymbol{q}^n, \boldsymbol{k}^n, \boldsymbol{v}^n$ are all vectors of dimension $d \times 1$.

(2) *Self Attention:* For the $i^{\text{th}}$ token, the key and value vectors of all tokens up to and including token $i$ are stacked into matrices $K^{n,i}$ and $V^{n,i}$ of dimension $d \times i$, and the self-attention is computed as,

$$\boldsymbol{y}^n = V^{n,i}\text{softmax}\left(\frac{{K^{n,i}}^\top \boldsymbol{q}^n}{\sqrt{d}}\right), \tag{1}$$

where $\boldsymbol{y}^n \in \mathbb{R}^d$ is the resulting context vector. (1) represents a single attention head. In practice, Transformers employ multiple attention heads [30], meaning that multiple sets of $Q$, $K$, and $V$ vectors are used in parallel in a layer. The resulting outputs are then concatenated and passed to a Feed-Forward Network (FFN) described below. For simplicity, and without loss of generality, we focus on a single attention head in our theoretical analysis. Our results are also applicable to memory-efficient variants such as multi-query attention [27] and grouped-query attention [3].

(3) *Feed Forward Net (FFN):* The output of the self-attention sublayer is passed through an FFN with one hidden-layer. The dimensions of the weight matrices of the FFN, $W_{F_1}^n$ and $W_{F_2}^n$, are $d_{ff} \times d$ and $d_x \times d_{ff}$ respectively. The output of the FFN serves as the input to the $(n+1)^{\text{th}}$ layer.

The output of layer N is passed through a linear transform followed by a softmax operation. The probability distribution output by the softmax is used to sample the next token in the sequence.

## 2.2 Inference on a Transformer

An LLM inference system receives requests in the form of *prompts*, consisting of a variable length sequence of tokens $\theta_1, \theta_2, \ldots, \theta_{l^P}$, where $l^P$ denotes the *prompt length*. The inference task is to autoregressively sample *output tokens* $\theta_{l^P+1}, \theta_{l^P+2} \ldots, \theta_{l^P+l^D}$ from the Transformer until a `stop` token is sampled at an apriori unknown position $l^P + l^D + 1$, where $l^D$ denotes the *output length*. A central technique used to speed up LLM inference is KV caching, wherein the key and value vectors (KV vectors) that have been computed for a token $\theta_i$ are cached in GPU memory. This allows subsequent tokens to directly retrieve the relevant KV vectors during attention computation, avoiding recomputation and significantly improving inference efficiency.

The computation associated with completing an LLM inference request can be divided into two distinct phases,

(1) *Prefill-Phase*: Although prompt tokens are specified by the request and thus do not need to be sampled, the KV vectors for all prompt tokens do need to computed and cached for all layers since they are required to perform the self-attention computation for later tokens. The computation of the KV vectors corresponding to prompt tokens is called the prefill-phase. Since all the prompt tokens are available, the prefill-phase can be done in parallel for all the prompt tokens at once, or a *chunk* of tokens at a time. This is discussed further in Sections 3.1 and 3.2.
(2) *Decode-Phase*: The decode-phase denotes the computations involved in generating output tokens $\theta_{l^P+1}, \theta_{l^P+2}, \ldots, \theta_{l^P+l^D}$. The KV vectors computed in this phase are also cached in memory for use in the self-attention computation of future tokens. Since this is the token generation phase, it is autoregressive in nature unlike the prefill-phase, and thus due to dependencies cannot be parallelized across tokens of the request.

See Fig. 2 for a visualization of the two phases of LLM inference.

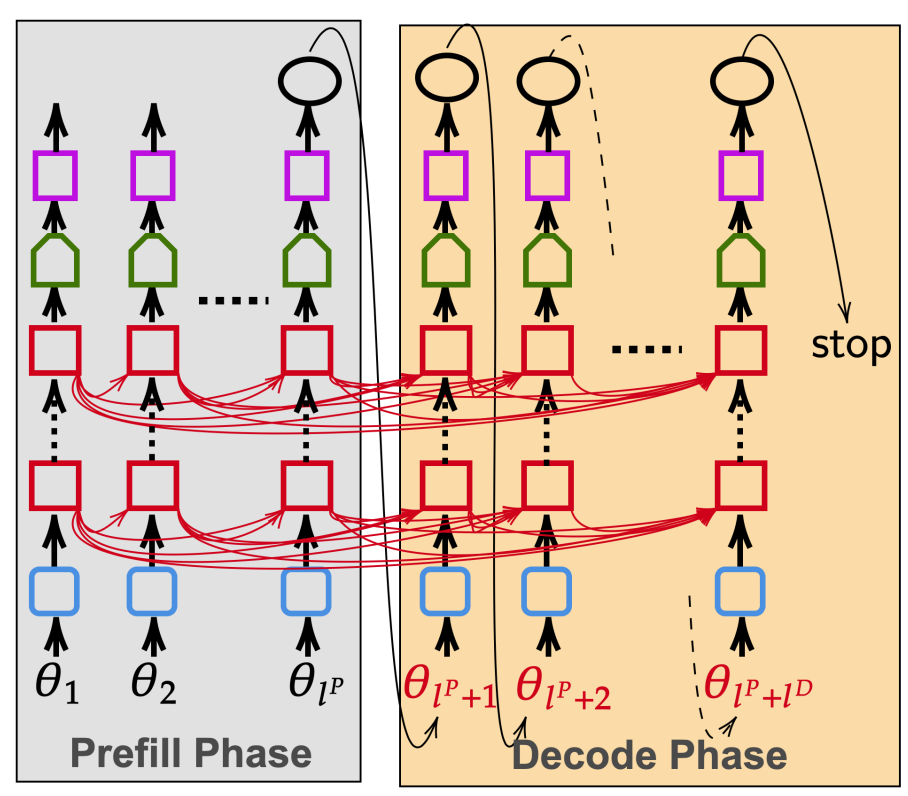

Fig. 2. Prefill and decode phases of a request during inference on a Transformer.

## 2.3 Computation on a GPU

A significant fraction of computations associated with LLM inference and, more generally, modern AI workloads involve matrix-matrix multiplication which can be sped up by specialized hardware. As such, modern GPUs have several components to perform matrix-matrix multiplication that is distinct from components used to perform general parallel computations. For concreteness, in this paper we focus our analysis on the terminology and convention used by NVIDIA TensorCore GPUs [25], while noting that GPUs produced by Intel[10], AMD[1] etc., have analogous components. From hereon, unless otherwise mentioned, by GPU we mean NVIDIA TensorCore GPU.

A GPU consists of a set of $s$ independent processors called Streaming Multiprocessors (SMs). An SM in turn consists of a number of fundamental computational units, which can be categorized into two categories: 1) CUDA cores which are general purpose parallel computation units, and 2) Tensor Cores that are specifically designed to perform (generalized) matrix-matrix multiplications.

Two types of computational tasks on a GPU are especially relevant in the context of LLM inference, Generalized Matrix-Matrix Multiplication (GeMM) and Generalized Matrix Vector Multiplication (GeMV), which are detailed below.

***GeMM.*** Let $A \in \mathbb{R}^{d_{\text{row}} \times d_{\text{red}}}$, $B \in \mathbb{R}^{d_{\text{red}} \times d_{\text{col}}}$, $C \in \mathbb{R}^{d_{\text{row}} \times d_{\text{col}}}$ and let $\alpha, \beta \in \mathbb{R}$; the GeMM operation computes $C \longleftarrow \alpha AB + \beta C$. In this formulation $d_{\text{row}}$ and $d_{\text{col}}$ denote the row and column dimensions of the output matrix $C$, while $d_{\text{red}}$ is the *reduction dimension*, corresponding to the shared inner dimension of input matrices $A$ and $B$.

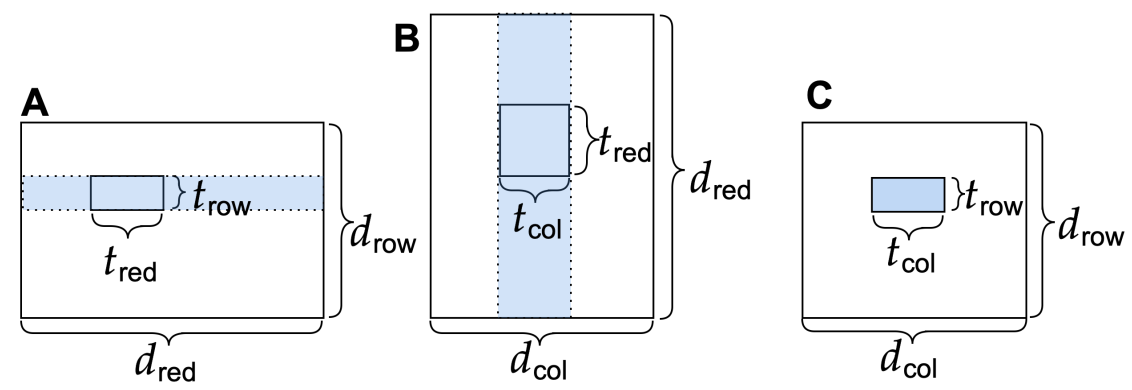

Fig. 3. Computation of a tile in a GeMM, $C \longleftarrow \alpha AB + \beta C$, requires a sweep of a row of tiles of $A$ and a column of tiles of $B$.

The GPU GeMM algorithm [22] partitions the output matrix $C$ into tiles of size $t_{\text{row}} \times t_{\text{col}}$ (padding zeros if necessary to fit dimensions). The computation of a tile of $C$ is assigned to an SM. The input matrices $A$ and $B$ are tiled into tiles of dimensions $t_{\text{row}} \times t_{\text{red}}$ and $t_{\text{red}} \times t_{\text{col}}$. Here, we shall refer to $t_{\text{red}}$ as the reduction tile dimension. Each SM loads the corresponding input tiles into local memory to compute its assigned output tile. See Fig. 3 for a visual representation of this process.

Tensor Cores can only be used if the output tile dimensions belong to a particular GPU-dependent set, $\mathcal{T}_{\text{out}}$ and the reduction tile dimension belongs to a set $\mathcal{T}_{\text{red}}$. Tile dimensions in $\mathcal{T}_{\text{out}}$ and $\mathcal{T}_{\text{red}}$ are always powers of 2. For instance, on the A100 GPU, $\mathcal{T}_{\text{out}}^{A100} = \{(256, 128), (128, 256), (128, 128),$ $(256, 64), (64, 256), (128, 64), (64, 128), (64, 64)\}$, and typically $\mathcal{T}_{\text{red}} = \{32, 64\}$. Moreover, Tensor Cores provide a substantially higher FLOP/s rate for GeMM with any of the allowed tile configurations as compared to CUDA cores. Therefore it is most often beneficial to pad the input and output matrices with zeros in order to partition it into tiles that fit in $\mathcal{T}_{\text{out}}$ and $\mathcal{T}_{\text{red}}$, even though it leads to redundant (zero-padding) computation. When zero padding is used, the FLOP/s rate only counts the useful computations and excludes redundant computations, i.e., the theoretical peak is computed using the original matrix sizes, not the padded sizes. For a detailed discussion on the characteristics of matrix sizes and FLOP/s rate of GeMMs, refer [22].

Suppose the GeMM computation uses output tile dimensions $t_{\text{row}} \times t_{\text{col}}$ and reduction tile dimension $t_{\text{red}}$. The output matrix $C$ then contains $\lceil d_{\text{row}}/t_{\text{row}} \rceil \cdot \lceil d_{\text{col}}/t_{\text{col}} \rceil$ tiles, and each output tile requires $\lceil d_{\text{red}}/t_{\text{red}} \rceil$ tile-pairs of $A$ and $B$ to be multiplied (ref. Fig. 3). Let $\mu(t_{\text{row}}, t_{\text{col}}, t_{\text{red}})$ denote the speed of computing the GeMM of a tile-pair with dimensions $t_{\text{row}} \times t_{\text{red}}$ and $t_{\text{red}} \times t_{\text{col}}$, and recall that $s$ is the number of SMs on the GPU. Then, we model the total time to complete the GeMM as

$$T_{\text{GeMM}}(d_{\text{row}}, d_{\text{col}}, d_{\text{red}}, t_{\text{row}}, t_{\text{col}}, t_{\text{red}}) = \frac{1}{s\mu(t_{\text{row}}, t_{\text{col}}, t_{\text{red}})} \left\lceil \frac{d_{\text{row}}}{t_{\text{row}}} \right\rceil \left\lceil \frac{d_{\text{col}}}{t_{\text{col}}} \right\rceil \left\lceil \frac{d_{\text{red}}}{t_{\text{red}}} \right\rceil. \tag{2}$$

Here, $1/s$ accounts for the parallelism provided by the $s$ SMs, $1/\mu(t_{\text{row}}, t_{\text{col}}, t_{\text{red}})$ captures the effective time to compute one tile-pair multiplication, and the remaining terms count the total number of tile-pair multiplications needed.

Assumption 1. *There exists output tile dimensions $(t^*_{\text{row}}, t^*_{\text{col}}) \in \mathcal{T}_{\text{out}}$ and a reduction tile size $t^*_{\text{red}} \in \mathcal{T}_{\text{red}}$ such that if a GeMM's matrices are a) sufficiently large, i.e., can fully utilize the GPU's compute resources, and b) are perfectly tiled i.e., there is no padding overhead, then the GPU achieves its maximal FLOP/s rate.*

Assumption 1 is well supported by empirical studies of GeMM performance on modern GPUs [22]. Intuitively, when matrices are sufficiently large and partition exactly into the tile size $t^*_{\text{row}} \times t^*_{\text{col}}$ along the output dimensions and $t^*_{\text{red}}$ along the reduction dimension, the GPU can fully utilize its SMs by creating a large number of independent tile computations. These output tile dimensions $t^*_{\text{row}} \times t^*_{\text{col}}$ are typically the largest in $\mathcal{T}_{\text{out}}$ which leads to higher arithmetic intensity–the ratio of floating-point operations to memory accesses–thereby enabling more efficient use of memory bandwidth and allowing the GPU to operate at its peak throughput. For instance, in the A100 GPU, $t^*_{\text{row}} \times t^*_{\text{col}}$ is either $128 \times 256$ or $256 \times 128$ and $t^*_{\text{red}}$ is 32.

***GeMV.*** Let $A \in \mathbb{R}^{d_{\text{row}} \times d_{\text{col}}}$, $\boldsymbol{x} \in \mathbb{R}^{d_{\text{col}}}$, and $\boldsymbol{y} \in \mathbb{R}^{d_{\text{row}}}$. Then, for scalars $\alpha, \beta$, the GeMV operation computes $\boldsymbol{y} \leftarrow \alpha A\boldsymbol{x} + \beta\boldsymbol{y}$. The GeMV is also implemented by tiling the output vector $\boldsymbol{y}$ into sub-vectors of size $t_{\text{row}}$, which are assigned to individual SMs and computed using CUDA cores. Correspondingly, the matrix $A$ is partitioned into tiles of size $t_{\text{row}} \times t_{\text{col}}$, and the input vector $\boldsymbol{x}$ into tiles of size $t_{\text{col}}$. Let $\mu(t_{\text{row}}, t_{\text{col}})$ denote the speed of computing a single tile-pair GeMV. Then the total time to complete the GeMV is modeled as

$$T_{\text{GeMV}}(d_{\text{row}}, d_{\text{col}}, t_{\text{row}}, t_{\text{col}}) = \frac{1}{s\mu(t_{\text{row}}, t_{\text{col}})} \left\lceil \frac{d_{\text{row}}}{t_{\text{row}}} \right\rceil \left\lceil \frac{d_{\text{col}}}{t_{\text{col}}} \right\rceil. \tag{3}$$

Because GeMV performs fewer floating-point operations per byte of memory traffic than GeMM, its arithmetic intensity is lower and thus it is less efficient in terms of memory bandwidth utilization. This motivates the following assumption.

Assumption 2. *Let $d_{\text{row}}, d_{\text{col}}, d_{\text{red}} \in \mathbb{N}$. Let $A \in \mathbb{R}^{d_{\text{row}} \times d_{\text{red}}}$. Let, $\boldsymbol{x}(1), \ldots, \boldsymbol{x}(d_{\text{col}}) \in \mathbb{R}^{d_{\text{red}} \times 1}$ and $\boldsymbol{y}(1), \ldots, \boldsymbol{y}(d_{\text{col}}) \in \mathbb{R}^{d_{\text{row}} \times 1}$ be $2d_{\text{col}}$ vectors. Denote the respective stacked vectors as, $X = [\boldsymbol{x}(1)\, \boldsymbol{x}(2)\, \ldots, \boldsymbol{x}(d_{\text{col}})]$ and $Y = [\boldsymbol{y}(1)\, \boldsymbol{y}(2)\, \ldots\, \boldsymbol{y}(d_{\text{col}})]$. Let $\alpha, \beta \in \mathbb{R}$. We will assume, that the time required to compute the set of GeMVs, $\boldsymbol{y}(i) \leftarrow \alpha A\boldsymbol{x}(i) + \beta\boldsymbol{y}(i)$ for all $i \in \{1, \ldots, d_{\text{col}}\}$, is greater than the time required to compute the GeMM, $Y \leftarrow \alpha AX + \beta Y$.*

## 3 LLM Inference System Model

We shall consider an LLM inference system consisting of $r$ *inference nodes* coordinated by a central *resource planner* as shown in Fig. 4 which function as follows.

*Inference node:* An inference node is the basic computational unit responsible for performing LLM inference. For large models such as GPT-3, PaLM, etc., inference is typically performed using multiple GPUs due to memory constraints. However, all GPUs within a node participate in a fixed execution pipeline using tensor-parallelism[21] or pipeline-parallelism[9]. Without loss of generality, it is convenient to model an inference node as operating on a single GPU.

The primary decision process at an inference node lies in its *LLM inference scheduler* which decides the order in which to schedule tasks on the node's GPU. The design of LLM inference schedulers is the focus of this work, and, we describe a general framework to study them in Section 3.1, after a brief description of the resource planner below.

*Resource planner:* The resource planner receives incoming requests in the form of prompts and assigns each one to an inference node, where the request joins a queue and waits to be served. The planner continuously monitors the status of all inference nodes—tracking metrics such as the number of pending requests, request latencies, and memory usage (including risks of overflows). Based on this information, it may choose to reassign waiting requests or even migrate partially processed requests to other nodes. When a partially completed request is migrated, the source node must transfer the partially computed KV cache to the destination node, which might incur a cost such as the time required to complete the transfer.

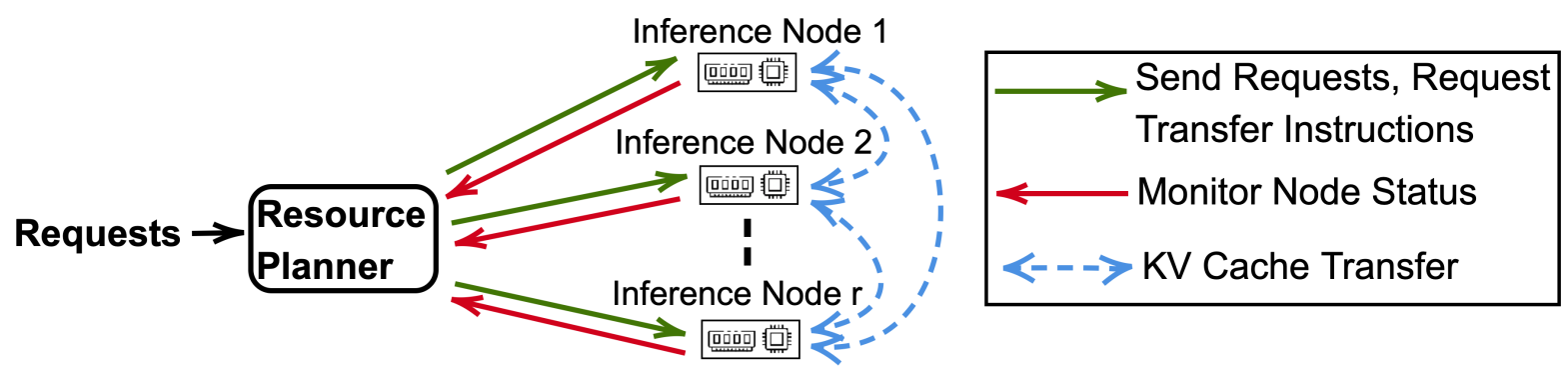

Fig. 4. An LLM inference system.

A resource planner in some systems, such as in NVIDIA Dynamo[4], has the ability to turn off some inference nodes when they are being underutilized, and turn them on when the system is congested. This feature is especially useful towards achieving energy efficiency when the request arrival pattern is heterogeneous and may vary through time. We focus on stationary request arrivals in our work for analytical tractability, and do not study policies to turn nodes on or off.

### 3.1 A General Framework for LLM Inference Schedulers

We consider a class of schedulers that schedule batches of request iterations non-preemptively that captures a range of existing schedulers in the literature. In this context, the meaning of an iteration of a request depends on whether it is in the prefill-phase or the decode-phase, as defined below.

Definition 1 (Prefill-Iteration). *For a request $R$, a chunk-size $c$, and starting token index $i$, a prefill-iteration, denoted $PI(R, i, c)$, involves computing the KV vectors for each layer corresponding to prompt token indices in the range $i$ to $i + c - 1$ (both inclusive). Once the iteration is complete, the starting token index is updates to $i + c$.*

In order to schedule $PI(R, i, c)$ on the GPU, the KV vectors for tokens up to index $i - 1$ must have been previously computed and cached in GPU memory, and $i + c - 1$ should not exceed the prompt length. Moreover, if $i + c - 1$ corresponds to the last prompt token, then the prefill-iteration generates the first output token.

Definition 2 (Decode-Iteration). *An iteration of a request $R$ in the decode-phase, denoted $DI(R, i)$, corresponds to the generation of the next output token which is at index $i + 1$.*

In order to schedule $DI(R, i)$, all the KV vectors for tokens up to index $i - 1$ must be previously computed and cached in GPU memory. Moreover, index $i$ should not correspond to a prompt token, and should not correspond to a token that follows the `stop` token.

Let $\mathcal{P}_m$ and $\mathcal{D}_m$ be the sets of requests that are in their prefill-phase and decode-phase respectively just prior to the construction of batch $m$. Let $\mathcal{B}_m$ denote the set of iterations to be scheduled in batch $m$. Initialize $\mathcal{B}_m = \emptyset$. A scheduler selects a subset $\mathcal{S}_m^P$ of $b_m^P$ prefill-phase requests from $\mathcal{P}_m$ and subset $\mathcal{S}_m^D$ of $b_m^D$ decode-phase requests from $\mathcal{D}_m$ and constructs the batch as follows: 1) for each selected prefill-phase request $R_j \in \mathcal{S}_m^P$ with starting token index $i_j$, it chooses a chunk size $c_j$ and adds $PI(R_j, i_j, c_j)$ to $\mathcal{B}_m$; 2) for each selected decode-phase request $R_j \in \mathcal{S}_m^D$ with the latest token index $i_j$, it adds $DI(R_j, i_j)$ to $\mathcal{B}_m$. At most one decode-iteration can be scheduled per request in a batch. Then, it schedules $\mathcal{B}_m$ on the GPU. Note that all the iterations within the batch may be processed in parallel by the GPU. A description of some existing LLM schedulers expressed in this framework is provided in Appendix E.

Next, we discuss how computations corresponding to a batch are performed on the GPU.

### 3.2 LLM Inference Computation on the GPU

We split the operations involved into 3 categories.

*3.2.1 Linear transformations:* This category includes the QKV projections in each layer, the linear computations in the FFN sub-layer of each layer and also the final linear projection. In all computations in this category, a weight matrix of the transformer multiplies a token feature vector. In particular, these computations involve the features of the present token, and do not involve any previous tokens in the sequence.

In line with Assumption 2, this operation may be scheduled efficiently as a GeMM instead of multiple GeMVs as follows. Let $W$ denote the weight matrix of the transformer. Let $\tau_m$ denote the *token-count* of the $m^{\text{th}}$ batch, which is, $\tau_m = b_m^D + \sum_{j:R_j \in \mathcal{S}_m^P} c_j$. Stack all the $\tau_m$ corresponding feature vectors into a matrix $Z^{in}$ and compute $Z^{out} = WZ^{in}$. Let $(t_{\text{row}}, t_{\text{col}}) \in \mathcal{T}_{\text{out}}$ and $t_{\text{red}} \in \mathcal{T}_{\text{red}}$ denote the output tile dimensions and reduction tile dimension used respectively for all linear transformations. For simplicity, we use a common tile configuration across transformations, as Assumption 1 implies a single tile configuration can achieve optimal performance across different GeMM dimensions.

Consider the QKV projection in layer $n$. The dimensions of $W_Q^n$, $W_K^n$ and $W_V^n$ are $d \times d_x$. Then, $X^n$, which is the stack of $\tau_m$ number of $\boldsymbol{x}^n$'s, has a dimension of $d_x \times \tau_m$. And the 3 output matrices, which are the stacks of $\boldsymbol{q}^n$'s, $\boldsymbol{k}^n$'s and $\boldsymbol{v}^n$'s respectively of tokens in the batch, have a dimension of $d \times \tau_m$ each. Since the model dimensions are designed with hardware in mind, $d, d_x$ (as well as other model dimensions) are divisible by the tile dimensions. Therefore, according to (2), the time required to complete the 3 GeMMs is, $\frac{3}{s\mu(t_{\text{row}},t_{\text{col}},t_{\text{red}})} \frac{d_x}{t_{\text{red}}} \frac{d}{t_{\text{row}}} \left\lceil \frac{\tau_m}{t_{\text{col}}} \right\rceil$. The two FFN sub-layer matrices $W_{F_1}^n$ and $W_{F_2}^n$ have dimensions $d_{\text{ff}} \times d$ and $d_x \times d_{\text{ff}}$, respectively. Denote the dimension of the final linear projection matrix as $d_{\text{out}} \times d_x$. Recalling that there are $N$ layers in the transformer, the amount of time required to compute the linear transformations for a batch with token count $\tau_m$ is,

$$\frac{1}{s\mu(t_{\text{row}}, t_{\text{col}}, t_{\text{red}})} \left( 3N \frac{d_x}{t_{\text{red}}} \frac{d}{t_{\text{row}}} + N \frac{d}{t_{\text{red}}} \frac{d_{\text{ff}}}{t_{\text{row}}} + N \frac{d_{\text{ff}}}{t_{\text{red}}} \frac{d_x}{t_{\text{row}}} + \frac{d_x}{t_{\text{red}}} \frac{d_{\text{out}}}{t_{\text{row}}} \right) \left\lceil \frac{\tau_m}{t_{\text{col}}} \right\rceil$$

In the above formula, the scheduler can only modify the token count of the batch, but not any of the transformer's parameters. So, we can abstract out fixed components, and express the time required to complete the linear transformation computations for the tokens in a batch as,

$$\frac{1}{\mu_{\text{Lin}}(t_{\text{row}}, t_{\text{col}}, t_{\text{red}})} \left\lceil \frac{\tau_m}{t_{\text{col}}} \right\rceil. \tag{4}$$

*3.2.2 Self-attention:* This computation has different characteristics in decode and prefill iterations.

The self-attention for a decode-iteration of a request consists of performing GeMV operations which cannot be scheduled as a GeMM using batching across requests because each request has a unique KV cache. Consider a decode-iteration $DI(R_j, i_j)$ in the batch. Its self-attention computation involves 2 GeMV's: one with a matrix of size $d \times i_j$ and a vector of length $i_j$, and another with a matrix of size $i_j \times d$ and a vector of length $d$. Let $(t_{\text{row}}, t_{\text{col}})$ be the optimal tile configuration for the GeMV. Then, from (3), the time required to perform Self-Attenion at layer $n$ for this decode-iteration is,

$$\frac{1}{\mu(t_{\text{row}}, t_{\text{col}})} \left( \frac{d}{t_{\text{col}}} \left\lceil \frac{i_j}{t_{\text{row}}} \right\rceil + \left\lceil \frac{i_j}{t_{\text{col}}} \right\rceil \frac{d}{t_{\text{row}}} \right) \triangleq T^{D,SA}(i_j). \tag{5}$$

Above, we use $T^{D,SA}(\cdot)$ to encapsulate the fixed components of the self-attention computation time.

On the other hand, since prefill-iterations may be scheduled for a chunk of tokens at a time, their self-attention may be computed as a GeMM as follows. Consider a prefill-iteration $PI(R_j, i_j, c_j)$ at layer $n$ of the Transformer. Stack the $i_j + c_j - 1$ each of key and value vectors of dimension $d$ of the first $i_j + c_j - 1$ tokens into matrices $K^n$ and $V^n$. Stack the $c_j$ query vectors of dimension $d$ corresponding to the current chunk into the matrix $Q^n$. Let $M$ be a masking matrix of dimension $(i_j + c_j - 1) \times c_j$, where each entry is 1 except in indices $(k_1, k_2)$ satisfying $k_1 > i$ and $k_2 \le k_1 - i$, where it is set to 0. The masking matrix allows a token to only attend to previous tokens in the prompt sequence. Then, the stack of output vectors of the self-attention may be computed as, $Y^n = V^n \text{softmax}\left(\frac{K^{n\top} Q^n}{\sqrt{d}} \odot M\right)$, where $\odot$ denotes an element-wise product. Observe that this requires: 1) a GeMM between matrices of dimensions $(i_j+c_j-1) \times d$ and $d \times c_j$, 2) element-wise multiplication with the mask $M$ followed by a non-linearity (softmax), and 3) another GeMM between $d \times (i_j+c_j-1)$ and $(i_j+c_j-1) \times c_j$ matrices. Let $(t_{\text{row}}, t_{\text{col}}) \in \mathcal{T}_{\text{out}}$ and $t_{\text{red}} \in \mathcal{T}_{\text{red}}$ denote the output tile dimensions and reduction tile dimension used respectively for both the GeMMs. We assume the cost of masking and softmax is negligible for simplicity. Then, according to (2), the time required to compute the self-attention across all $N$ layers is:

$$\frac{N}{s\mu(t_{\text{row}}, t_{\text{col}}, t_{\text{red}})} \left( \left\lceil \frac{i_j + c_j - 1}{t_{\text{row}}} \right\rceil \left\lceil \frac{c_j}{t_{\text{col}}} \right\rceil \frac{d}{t_{\text{red}}} + \frac{d}{t_{\text{row}}} \left\lceil \frac{c_j}{t_{\text{col}}} \right\rceil \left\lceil \frac{i_j + c_j - 1}{t_{\text{red}}} \right\rceil \right). \tag{6}$$

Remark 1. *For simplicity, we describe the computation of self-attention as two successive GeMM's interleaved with non-linear operations. In practice, systems usually use FlashAttention [5]-a hardware-efficient algorithm which minimizes the number of memory transfers by fusing operations and re-ordering computation using tiling. Our approach and analysis can be adapted for FlashAttention with appropriate modifications to* (6). *We leave this to future work.*

*3.2.3 Non-Linear Operations:* Each SM in a GPU has a *Special-Function-Unit* that efficiently computes non-linear functions. Transformers include non-linear operations such as ReLU, softmax, etc., which—like Linear Transformations—operate only on the tokens being processed in the batch. The time to compute all non-linear operations for a single token is denoted by $\frac{1}{\mu_{\text{nLin}}}$. Thus, for a batch containing $\tau_m$ tokens, the total computation time is $\frac{1}{\mu_{\text{nLin}}}\tau_m$, where $\mu_{\text{nLin}}$ denotes the rate at which non-linear operations are performed per token.

**Batch execution time:** Recall that $\mathcal{B}_m$ denotes the $m^{\text{th}}$ batch with prefill-iterations from requests in $\mathcal{S}_m^P$ and decode-iterations from requests in $\mathcal{S}_m^D$. Then, summing up the computations times from (4), (5) and (6) for all the iterations, we express the batch computation time as,

$$\begin{aligned} T_B(\mathcal{B}_m) = {} & \frac{1}{\mu_{\text{Lin}}(t_{\text{row}}, t_{\text{col}}, t_{\text{red}})} \left\lceil \frac{\tau_m}{t_{\text{col}}} \right\rceil + \frac{1}{\mu_{\text{nLin}}} \tau_m + N \sum_{j:R_j \in \mathcal{S}_m^D} T^{D,SA}(i_j) \\ & + \frac{N}{s\mu(t_{\text{row}}, t_{\text{col}}, t_{\text{red}})} \sum_{j:R_j \in \mathcal{S}_m^P} \left( \left\lceil \frac{i_j + c_j - 1}{t_{\text{row}}} \right\rceil \left\lceil \frac{c_j}{t_{\text{col}}} \right\rceil \frac{d}{t_{\text{red}}} + \frac{d}{t_{\text{row}}} \left\lceil \frac{c_j}{t_{\text{col}}} \right\rceil \left\lceil \frac{i_j + c_j - 1}{t_{\text{red}}} \right\rceil \right). \end{aligned} \tag{7}$$

The ceiling function in the batch execution time formula captures the staircase-behaviour observed in practice. See [17, Figure 4].

Next, we present our main theoretical contribution: a simple resource planner and a scheduler that maximizes the throughput—measured in requests per second—under stationary request arrivals.

## 4 A Throughput Optimal Inference System

### 4.1 Throughput optimal resource planner

For simplicity, we assume that all $r$ Inference Nodes in the system are identical and are running the same Transformer model. We consider a simple *uniformly random resource planner*, which distributes each incoming request uniformly at random to one of the $r$ inference nodes.

The random-routing assumption represents a simplified setting. The multi-node model is included to reflect large-scale inference architectures, which typically involve multiple GPUs or servers. Within the homogeneous setup considered here, random routing suffices to achieve throughput optimality. In more realistic deployments—featuring heterogeneous nodes, bursty workloads, or latency-sensitive objectives—more sophisticated routing strategies may be required. This model thus serves as a foundational step toward analyzing such richer settings.

A general resource planner in our model (see Fig. 4) has the ability to monitor node statuses and transfer pending or partially processed requests from one node to another. While the simple resource planner above does neither of these, we show that, combined with our proposed scheduler in Section 4.2, it maximizes resource utilization and is thus throughput-optimal, even when the cost of transferring the KV cache of partially processed requests between nodes is 0. The ability to monitor node status and transfer requests becomes particularly useful in practical systems that aim to minimize delay or satisfy strict latency SLO constraints [4].

### 4.2 RAD: A throughput optimal scheduler

Our proposed Resource-Aware-Dynamic (RAD) Scheduler is shown in Algorithm 1. The time horizon of RAD can be partitioned into *cycles*, where in a cycle, RAD schedules a sequence of batches that start and complete up to $n$ requests, with $n$ being a parameter. Now, in addition to $n$, RAD also accepts the optimal output tile dimensions, $(t^*_{\text{row}}, t^*_{\text{col}}) \in \mathcal{T}_{\text{out}}$ and reduction tile dimension $t^*_{\text{red}} \in \mathcal{T}_{\text{red}}$ as in Assumption 1, for optimal GeMM computation on the GPU. The central design principle of RAD is two-fold.

*Optimal GeMM tiling*. RAD schedules batches such that all GeMM computations in the batch are optimally tiled. It does so by either scheduling $t^*_{\text{col}}$ decode-iterations in a batch, or, scheduling a prefill-iteration with chunk size equal to $\text{LCM}(t^*_{\text{row}}, t^*_{\text{col}}, t^*_{\text{red}})$, where LCM stands for least-common-multiple. This tiling strategy is violated only in rare cases—specifically, near the end of a cycle—as further explained in the description of Algorithm 1.

*Optimal resource allocation* via dynamic scheduling of prefill and decode. RAD dynamically prioritizes prefill or decode iterations based on the request arrival pattern and its characteristics. For instance, if the requests have very long prompts but short output lengths, it spends more time performing prefill iterations. On the other hand, if the requests have relatively shorter prompt lengths and longer output lengths, it spends more time performing decode iterations. In this context, prefill and decode phases may be viewed as two classes of tasks, and that RAD partitions the node's service capacity such that both prefill and decode tasks may be completed.

In addition to the above principles, RAD also manages GPU memory and prevents overflow by limiting the number of active requests —requests that currently store their KV cache in GPU memory— to $t^*_{\text{col}}$, the number required for optimal tiling. See Assumption 4 for further details.

At a high level, RAD alternates between two operational modes—Prefill Mode and Decode Mode—as illustrated in Figure 5. At the start of a cycle, RAD enters Prefill Mode, completing $t^*_{\text{col}}$ prefill-phase requests, after which the same number of decode-phase requests become active. It then transitions to Decode Mode, scheduling decode iterations for these active requests in optimally tiled batches. When one or more decode-phase requests complete, RAD switches back to Prefill Mode to process additional prefill-phase requests until there are again exactly $t^*_{\text{col}}$ active decode-phase requests, thereby re-enabling efficient decode batching. Once a sufficient number of requests have been completed in the current cycle, RAD finishes the remaining ones—even if their batches are not optimally tiled—before beginning the next cycle.

Next, we provide a detailed explanation of RAD as shown in Algorithm 1. Let $\mathcal{P}$ and $\mathcal{D}$ denote the set of requests in prefill and decode phases respectively. A variable, $requests_in_cycle$ keeps track of how many requests have started processing (i.e., have had an iteration scheduled) in the current cycle. During a cycle, we continue to update $\mathcal{P}$ with any requests that arrive. Then at the completion of a batch, the next batch is selected as follows. If there are no requests in $\mathcal{P}$ or $\mathcal{Q}$, the scheduler simply waits until a new arrival happens (lines 5). Otherwise, 1) *If there are enough decode*

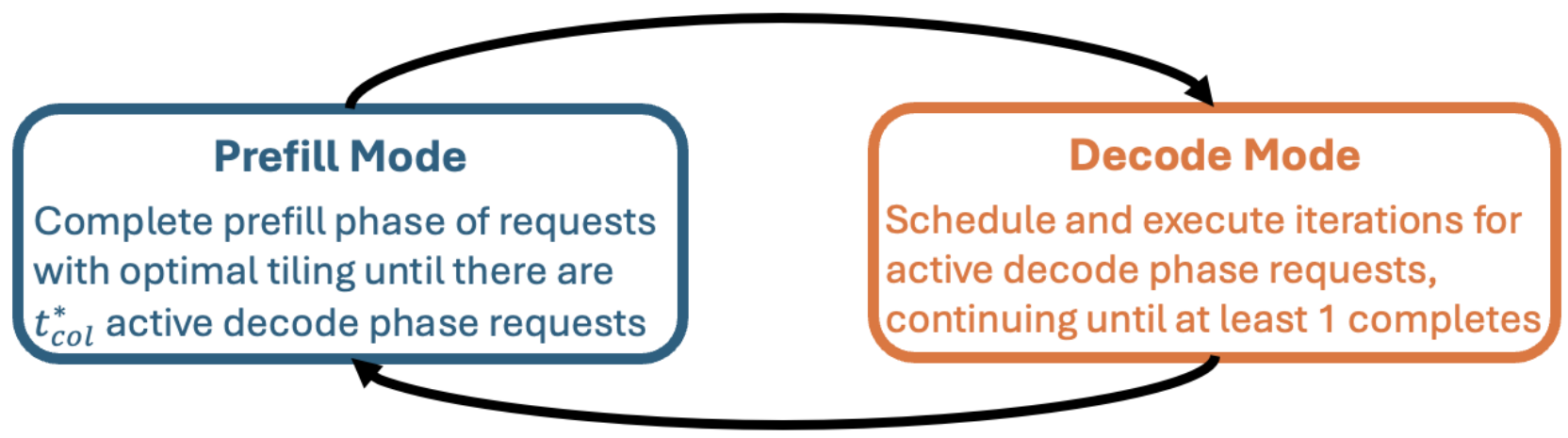

Fig. 5. High-level operation of the RAD scheduler. In Prefill Mode, the scheduler completes prefill iterations with optimal tiling until there are $t^*_{\text{col}}$ active decode-phase requests, at which point it switches to Decode Mode. In Decode Mode, it schedules decode iterations for these active requests until at least one completes, after which it switches back to Prefill Mode.

*phase requests, or there are no prefill phase requests, or n requests have started processing in the current cycle (Lines 6-11):* RAD selects one decode iteration per request in $\mathcal{D}$ and runs them together as a batch. After the GPU completes processing the batch, any requests that sampled the `stop` token are removed. If $\mathcal{D}$ is now empty, it signals the end of the current cycle, and *requests_in_cycle* is set to 0. 2) *Else (Lines 12-21)*: RAD chooses a request in a first-come-first-serve (FCFS) manner from $\mathcal{P}$ and schedules its prefill-iterations in chunks of size $\text{LCM}(t_{\text{row}}, t_{\text{col}}, t_{\text{red}})$, unless there are fewer tokens, until its prefill phase is complete. The request is then moved to $\mathcal{D}$, and the number *requests_in_cycle* is updated.

**Algorithm 1** Resource Aware Dynamic (RAD) Scheduler

**Require:** Parameter $n$ `// Max number of prefills per cycle`
**Require:** Output tile size $(t^*_{\text{row}}, t^*_{\text{col}})$ and reduction tile dimension $t^*_{\text{red}}$ `// For optimal tiling`
1: $\mathcal{P} \leftarrow \emptyset, \mathcal{D} \leftarrow \emptyset$ `// Set of requests in prefill and decode phase`
2: $requests_in_cycle \leftarrow 0$
3: **while** True **do**
4:     Update $\mathcal{P}$ based on new arrivals
5:     **if** $\mathcal{P} = \emptyset$ **and** $\mathcal{D} = \emptyset$ **then continue** `// Do nothing`
6:     **else if** $|\mathcal{D}| = t^*_{\text{col}}$ **or** $\mathcal{P} = \emptyset$ **or** $requests_in_cycle = n$ **then**
7:         Schedule a DI with appropriate token index for each request in $\mathcal{D}$ in a batch
8:         Remove requests that sampled the `stop` token from $\mathcal{D}$
9:         **If** $\mathcal{D} = \emptyset$ **then** $requests_in_cycle \leftarrow 0$ `// End of cycle`
10:     **else**
11:         Choose a request $R$ first-come-first-serve from $\mathcal{P}$ and set $i \leftarrow 1$
12:         **while** prefill phase of the chosen request is incomplete **do**
13:             $c = \min\{t^*_{\text{lcm}}, L^P(R) - i\}$ `//` $t^*_{\text{lcm}} = \text{LCM}(t^*_{\text{row}}, t^*_{\text{col}}, t^*_{\text{red}})$
14:             Schedule $PI(R, i, c)$ in a batch
15:             $i \leftarrow i + c$
16:         **end while**
17:         Move the request from $\mathcal{P}$ to $\mathcal{D}$
18:         $requests_in_cycle \leftarrow requests_in_cycle + 1$
19:     **end if**
20: **end while**

### 4.3 Throughput Optimality Characterization

The following theorem characterizes the maximum arrival rate under which any routing and scheduling policy could be stable.

Theorem 1 (Upper Bound). *Consider a system with a resource planner and $r$ identical inference nodes. Let requests arrive to the system according to a Poisson-Point-Process (PPP) of rate $\lambda$. Let the $j^{th}$ request arrival, $R_j$, have prompt and output lengths, $L_j^P$ and $L_j^D$, sampled according to a bounded joint-distribution $p_{L^P,L^D}$, i.i.d., across requests.*

*Let $(t^*_{\text{row}}, t^*_{\text{col}}) \in \mathcal{T}_{\text{out}}$ and $t^*_{\text{col}} \in \mathcal{T}_{\text{red}}$ be the optimal output tile dimensions and reduction tile dimension respectively from Assumption 1, and let, $t^*_{\text{lcm}} = \text{LCM}(t^*_{\text{row}}, t^*_{\text{col}}, t^*_{\text{red}})$. Define,*

$$\bar{T}^R = \frac{1}{\mu_{Lin}(t^*_{\text{row}}, t^*_{\text{col}}, t^*_{\text{red}})} \frac{\mathbb{E}\left[L^P + L^D\right]}{t^*_{\text{col}}} + \frac{\mathbb{E}\left[L^P + L^D\right]}{\mu_{nLin}} + N\mathbb{E}\left[\sum_{i=1}^{L^D} T^{D,SA}(L^P + i)\right]$$
$$+ \frac{Nd}{s\mu(t^*_{\text{row}}, t^*_{\text{col}}, t^*_{\text{red}})t^*_{\text{row}}t^*_{\text{col}}t^*_{\text{red}}} \mathbb{E}\left[L^P(L^P + t^*_{\text{lcm}})\right].$$

*Denote $Q(t)$ to be the number of pending requests in the system at time $t$, i.e., those that have arrived but not yet completed service. Under Assumptions 1 and 2, if $\lambda \bar{T}^R > r$, there exists an $\alpha > 0$ such that for any resource planner and scheduler, $\lim_{t\to\infty} \frac{Q(t)}{t} > \alpha$ almost surely.*

The proof of Theorem 1 can be found in Appendix B. Intuitively, the argument proceeds by showing that, in the best case, an inference node may be viewed as a single server system with expected service time $\bar{T}^R$. Then, instability may be shown to occur when the load on the system exceeds its service capacity.

Next, we proceed to analyze the range of arrival rates under which the RAD scheduler stabilizes the system. First, we make two assumptions.

Assumption 3. *Let $(t^*_{\text{row}}, t^*_{\text{col}})$ and $t^*_{\text{red}}$ be output tile dimensions and reduction tile dimension used by RAD. Then, prompt lengths $L_P \sim p_{L^P}$ are multiples of $\text{LCM}(t^*_{\text{row}}, t^*_{\text{col}}, t^*_{\text{row}})$ almost surely. Finally, assume the prompt and output lengths are upper bounded by $l^{P,\max}$ and $l^{D,\max}$ respectively.*

The upper bound on the prompt and output lengths is natural since LLMs only support bounded context lengths. If the prompt length is not a multiple of $\text{LCM}(t^*_{\text{row}}, t^*_{\text{col}}, t^*_{\text{row}})$, then any scheduler may have to complete a part of the prefill-phase with non-optimal tiling. We make this assumption to simplify our scheduler and its analysis.

Assumption 4. *Let $(t^*_{\text{row}}, t^*_{\text{col}})$ be the output tile dimensions used by RAD. Each inference node is provisioned with sufficient GPU memory to store the KV cache for $t^*_{\text{col}}$ concurrent requests, each with prompt length $l^{P,\max}$ and output length $l^{D,\max}$, where $l^{P,\max}$ and $l^{D,\max}$ are defined in Assumption 3.*

This assumption ensures that sufficient GPU memory is available to accommodate the worst-case KV cache requirements needed by the RAD scheduler. The above assumption is also a mild one because $t^*_{\text{col}}$ is typically as small as 64 or 128. Modern LLM inference nodes that use paged-attention [15] are able to accommodate the KV-cache of these many requests in their GPU memory.

Next, we state our result which characterizes the loads that the RAD scheduler combined with a simple resource planner can stabilize.

Theorem 2 (Throughput Optimality of RAD). *Consider a system with a uniformly random resource planner, and $r$ identical Inference Nodes running the RAD scheduler. Let requests arrive according to a Poisson Point Process (PPP) of rate $\lambda$. Let the $j^{th}$ request arrival, $R_j$, have prompt and output lengths, $L_j^P$ and $L_j^D$, sampled according to a joint-distribution $p_{L^P,L^D}$, i.i.d., across requests.*

*Suppose Assumptions 1, 2, 3, and 4 hold. Let $\bar{T}^R$ be as defined in Theorem 1. Define the worst-case time to complete a request as:*

$$T^{\max} \triangleq \max_{(t_{\text{row}},t_{\text{col}})\in\mathcal{T}_{\text{out}},\, t_{\text{red}}\in\mathcal{T}_{\text{red}}} \left[ \frac{l^{P,\max}+l^{D,\max}}{\mu_{Lin}(t_{\text{row}},t_{\text{col}},t_{\text{red}})} + \frac{l^{P,\max}+l^{D,\max}}{\mu_{nLin}} + N \sum_{i=1}^{l^{P,\max}+l^{D,\max}} T^{D,SA}(i) \right].$$

*Suppose there exists $\varepsilon > 0$ such that $\lambda\bar{T}^R \le r(1-\varepsilon)$. Then, RAD with parameter $n$ satisfying $n > \frac{(t^*_{\text{col}}-1)\,T^{\max}}{\varepsilon\bar{T}^R}$, is stable in the following sense. Let $Q_{r'}(t)$ denote the number of pending requests at node $r'$ at time $t$. For every $r'$, $(Q_{r'}(t))_{t\ge 0}$ forms a positive-recurrent regenerative process[28].*

We outline the arguments used to prove the statement above. First we construct a discrete-time Markov chain (DTMC) for each node that tracks the number of requests at the node at the start of each cycle. We show that the DTMC is positive recurrent under the conditions of the Theorem. Further, we show that the expected time of a cycle is finite. This leads us to conclude that the number of pending requests $Q_{r'}(t)$ at any time $t$ at a node $r'$ forms a positive-recurrent regenerative process— it visits 0 in finite expected time. The proof can be found in Appendix C.

## 5 Practical Insights from Theory

In Section 4, we identified two key principles that make a system throughput optimal: 1) optimal GeMM tiling when there are sufficient requests in the system, 2) optimal resource allocation for prefill and decode iterations to account for variability and uncertainty of prompt and output lengths.

In practice, LLM inference systems also have to meet other latency-based SLO constraints, such as TTFT and TBT as explained in the Section 1. We believe that designing a system that simultaneously achieves optimal GeMM tiling and optimal resource allocation, while maintaining low TTFT and TBT is challenging. As an illustrative example, consider an instance of running the RAD scheduler where in every batch of $t^*_{\text{col}}$ decode-iterations, one or more requests generate their `stop` token. In this case, the RAD scheduler has to complete prefill-phases of an equal number of new requests in between every consecutive batch of decode-iterations. Since prefill-phases may take a while to complete because of long prompt lengths, all requests may experience high TBT. See Appendix D for a throughput-optimal scheduler with low TBT but which incurs high TTFT.

In real-world deployments, meeting both TTFT and TBT constraints often necessitates a tradeoff between optimal tiling and optimal resource allocation. We examine two state-of-the-art LLM inference paradigms—*single-node serving*, where both the prefill and decode phases of each request are executed on the same inference node, and *disaggregated serving*, where these phases are distributed across different nodes—and present a novel perspective on how these systems adopt complementary strategies to navigate this tradeoff.

*Single node serving maintains optimal resource allocation but sacrifices optimal GeMM tiling.* Consider, Sarathi-serve [2] a state-of-the-art single node serving scheduler. When enough requests are present, Sarathi-serve prioritizes filling a batch with available decode iterations, then adds prefill iterations such that the total token count equals $\tau_{\text{budget}}$, a multiple of $t^*_{\text{col}}$. This dynamic batching strategy is similar to the RAD scheduler's, and preserves optimal allocation across prefill and decode. However, since it does not enforce that prefill chunk sizes are multiples of $\text{LCM}(t^*_{\text{row}}, t^*_{\text{col}}, t^*_{\text{red}})$, it does not guarantee optimal tiling. Other schedulers, such as vLLM [15] and Orca [33], also fall into the category of single-node serving schedulers.

*Disaggregated serving, by contrast, maintains optimal GeMM tiling but sacrifices optimal resource allocation.* Distserve [36], a state-of-the-art disaggregated serving scheduler, statically partitions inference nodes into prefill-nodes and decode-nodes. Prefill-nodes exclusively handle prefill iterations, which allows them to, in most cases, schedule with optimal chunk sizes, helping minimize

TTFT. Completed prefill-phase requests are transferred to decode-nodes. Decode-nodes handle decode iterations and can batch $t^*_{\text{col}}$ decode requests when available, keeping TBT low. However, static partitioning may lead to resource imbalance: for instance, when prompt lengths are short but output lengths are long, decode-nodes may become overloaded while prefill-nodes remain underutilized.

To address practical needs, we next consider the design of a scheduler that serves heterogeneous request classes subject to latency constraints at a single node. Since single node serving systems dynamically allocate resources and are therefore more robust to heterogeneity in request arrivals, we focus on this paradigm in the sequel.

## 6 SLO Aware LLM Inference Scheduler

***Request classes.*** In real-world LLM serving systems, requests may come from different classes of user with heterogeneous latency SLOs. For example, paying users typically expect fast and smooth responses, especially during token generation, which requires stricter TBT deadlines. By contrast, free-tier users are generally more tolerant of delays and can be served with more relaxed TBT constraints. Managing these mixed latency requirements well is important to keep users satisfied while making efficient use of system resources.

Recall that we consider a class of schedulers that executes prefill-phase and decode-phase requests as sequences of prefill-iterations and decode-iterations, respectively, see Definitions 1 and 2.

***Motivation.*** The throughput-optimal RAD scheduler described in Section 4 focuses on maximizing throughput, but it does not consider latency SLOs during either the prefill or decode phases due to the challenges mentioned in Section 5. However, in practice, meeting these latency constraints is critical to deliver a good user experience.

Sarathi-Serve, the current state-of-the-art scheduler, addresses this by chunking long prefill-phase requests into smaller chunks and interleaving them with decode-phase requests in each batch. Each batch is constrained by a token budget—the maximum number of tokens it can process. Sarathi-Serve includes all active decode-phase requests from the previous batch in the current one and uses the remaining token budget to schedule prefill-iterations. However, it treats all decode-iterations of associated decode-phase requests as if they had the strictest TBT deadline, even when actual TBT deadlines vary across requests. While this conservative strategy ensures tail TBT latency is below some threshold, it can lead to inefficient use of batch capacity and does not address reducing TTFT for prefill-phase requests. To overcome this limitation, we propose the SLO-Aware LLM Inference (SLAI) Scheduler. SLAI tracks each decode-iteration's TBT deadline and delays its inclusion in a batch until necessary. This allows the scheduler to allocate more of the batch's token budget to prefill-iterations earlier, without missing TBT deadlines on decode-iterations. As we will show, this approach better aligns scheduling decisions with request-specific needs, resulting in lower median TTFT for prefill-phase requests while still meeting tail TBT latency constraints on decode-iterations.

***Key concepts and parameters.*** We begin by explaining how SLAI dynamically decides when to include a decode-iteration in a batch. The key idea in this process is the *last schedulable time*, which determines when a decode-iteration becomes critical and must be included to meet its latency target.

Let $\delta > 0$ be an offset parameter for SLAI that defines how early a decode-iteration should be considered critical. Consider the $i^{\text{th}}$ decode iteration for a given request $j$ which has an SLO requirement of $\text{TBT}_j$, SLAI notes the end time of the most recent batch in which its $(i-1)^{\text{th}}$ decode-iteration was included, denoted $e_{i-1,j}$. It also maintains the running average of batch execution times observed so far, denoted by $\bar{t}_{\text{batch}}$. We define its *last schedulable time* as:

$$C_{i,j} = e_{i-1,j} + \text{TBT}_j - \delta \cdot \bar{t}_{\text{batch}} \tag{8}$$

This is the latest wall-clock time by which decode-iteration $j$ must be included in a batch to meet its TBT deadline. When constructing batch $m$ at time $t_m$, the scheduler checks each active (in progress that currently occupies GPU memory) decode-phase request and labels its decode-iteration as *critical* if $t_m \geq C_{i,j}$; otherwise, it is considered as *non-critical* and can be deferred to a later batch.

Besides this dynamic prioritization, SLAI uses several key parameters to balance latency targets with efficient GPU use. The token budget $\tau^{\text{budget}}$ sets the maximum token count allowed in a batch, ensuring effective use of the GPU's compute resources without violating TBT SLOs. A cap on the number of active requests $\alpha$ limits how many active requests are allowed at once, helping prevent memory overflows for large models or long prompts. A decode limit $\beta$ restricts how many of decode-iterations can be included in a batch, avoiding long batch execution times due to too many decode-iterations in a batch. Finally, the offset parameter $\delta$ provides a safety margin for the last schedulable time computation to absorb variability in batch execution and reduce the risk of missing TBT SLOs. We will later discuss how these parameters interact and influence the behavior and performance of the scheduler. Next, we describe how the SLAI scheduler works.

***Batch construction.*** We now describe how a batch is constructed by the SLAI scheduler. At each decision point, the scheduler forms a batch from the set of active requests and new requests that have not yet been processed. While forming a batch, it must respect several system constraints: the token budget ($\tau^{\text{budget}}$), the cap on active requests ($\alpha$), and the decode request limit ($\beta$). The batch construction follows these steps:

(1) *Identify critical decode-iterations:* For each active decode-phase request, compute the last schedulable time for its decode-iteration. A decode-iteration is marked as *critical* if current time is past its last schedulable time; otherwise, it is marked as *non-critical*.
(2) *Add critical decode-iterations:* Include critical decode-iterations in the batch, in the increasing order of last schedulable time. This ensures that decode-iterations that are closest to their TBT deadline are scheduled first.
(3) *Add prefill-phase requests:* Next, add prefill-phase requests in a non-preemptive manner. Among these, requests that have already been scheduled at least once (i.e., active prefill-phase requests) are given a higher priority. If token budget and cap on number of active requests has not been exceeded, new prefill-phase requests are considered. We consider two possibilities for ordering the incoming requests: Shortest Prefill First (SPF) to reduce the average or median TTFT or First Come First Serve (FCFS) order to ensure fairness.
(4) *Add non-critical decode-iterations:* Finally, if there further token budget remains and number of decodes in the batch are less than the decode limit, include additional non-critical decode-iterations in increasing order of their last schedulable time.

Next, we discuss the impact of parameters other than the offset ($\delta$), which has already been covered in our earlier discussion of the scheduler.

## 6.1 Impact of different scheduler parameters

*6.1.1 Token budget ($\tau^{budget}$).* The token budget places an upper limit on the number of tokens that can be processed in a single batch—one token per decode-phase request and $c \geq 1$ tokens per prefill chunk. Fig. 6a shows how the choice of token budget $\tau^{\text{budget}}$ affects TTFT and batch-execution time. When $\tau^{\text{budget}}$ is *small*, a 2048-token prefill must be split into many small chunks. This leads to frequent kernel launches and synchronization, causing the GPU to spend more time idling. As a result, TTFT increases. Conversely, when $\tau^{\text{budget}}$ is *large*, the scheduler can process the entire request in fewer large batches. This improves GPU utilization but each batch takes longer to execute. If a decode-iteration is scheduled during such a long batch, it must wait, increasing the risk of violating its TBT constraint. The scheduler must thus choose a token budget $\tau^{\text{budget}}$ that balances

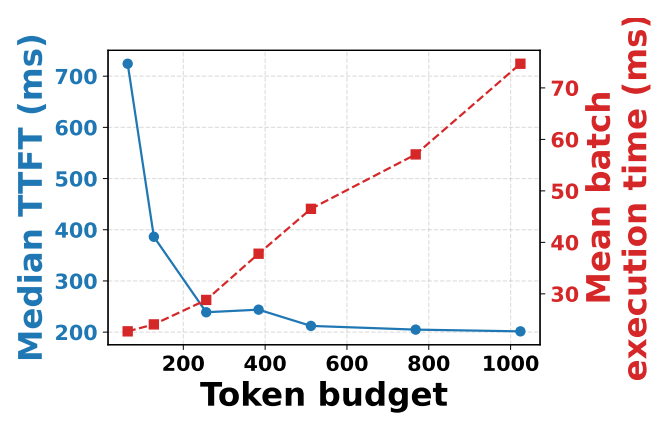

(a) Median TTFT and mean batch execution time for 100 independent requests (each with a 2048-token prefill and 1-token decode) as a function of the token budget.

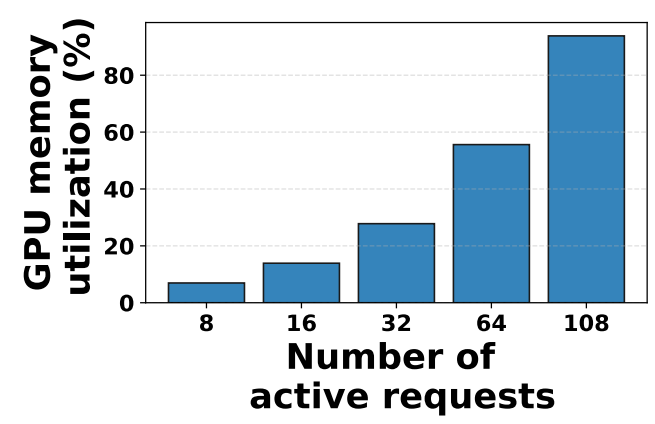

(b) GPU memory usage versus number of concurrent decode-phase requests (each with a 2048-token prefill and 1-token decode).

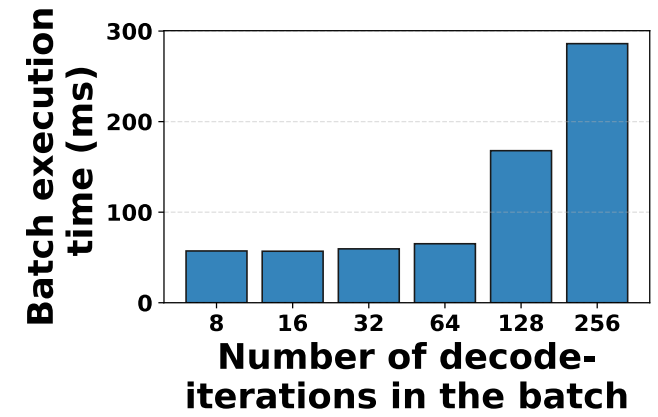

(c) Batch execution time for $N$ decode-phase requests (token position 513), co-scheduled with one prefill-only request to fully utilize the token budget of 512 tokens.

Fig. 6. Impact of token budget, concurrency, and batch composition on request execution latency and GPU memory usage for Mistral-7B on a single NVIDIA RTX ADA 6000 GPU.

efficiency and responsiveness. Batches should be large enough to use the GPU effectively, but not so long that they excessively delay latency-sensitive decode-phase requests. Choosing this value carefully is a key part of designing an effective scheduling policy.

*6.1.2 Cap on the number of active requests ($\alpha$).* Each request generates KV tensors that must be stored in GPU memory until the request is completed. Fig. 6b shows how GPU memory utilization grows with the number of active decode-phase requests, based on runs with $N$ concurrent decode-phase requests (each with a 2048-token prefill followed by a 1-token decode iteration) on Mistral-7B. As more requests become active, memory usage increases steadily.

When too many requests are active, the scheduler may need to evict KV tensors to make room for new ones. If a request's KV tensors are evicted, they must be recomputed before the request can resume decoding. This adds unnecessary delay and increases TTFT, even for requests that have not yet been scheduled. To avoid this, the scheduler should cap the number of active requests, ensuring that all necessary KV tensors can remain in memory without eviction. Doing so helps maintain low latency and avoids unnecessary recomputation overheads.

*6.1.3 Decode limit ($\beta$).* The decode limit sets an upper bound on how many decode-iterations can be included in a single batch. Fig. 6c shows how batch-execution time changes as the number of decode-iterations increases, while keeping the token budget fixed at 512. When only a few decode iterations are present, the batch finishes quickly. However, as more decode-iterations are added, the batch takes significantly longer to finish due to increased pressure on compute and memory resources. Each decode-iteration triggers self-attention computation, which involves GeMVs per request—an inefficient computation on GPU (see Section 3.2.2).

When many decode-phase requests are active, limiting the number of decode-iterations in each batch helps control latency. In order to meet strict TBT deadlines, the scheduler can cap the number of decode-iterations per batch, reducing the number of TBT violations and maintaining better responsiveness under load.

## 7 Experimental results

***Implementation.*** We built SLAI on top of the open-source implementations of Sarathi-serve [2] and vLLM [15]. The complete source code is available at: https://github.com/agrimUT/SLAI.

***Evaluation.*** We evaluate SLAI using the Mistral-7B [12] model (mistralai/Mistral-7B-Instruct-v0.2) and run all experiments on a single NVIDIA RTX 6000 Ada GPU with 48 GB of memory. The environment uses CUDA 12.1 and Python 3.10. For our workload, we use the `openchat_shareGPT4` [31] dataset, which contains multi-round conversations between users and ChatGPT4 [26]. Each round is treated as a separate request.

Our baseline is Sarathi-serve configured with FCFS ordering for prefill-phase requests, referred to as Sarathi-serve (FCFS), which represents the current state-of-the-art in LLM inference scheduling on a single node. We also evaluate a variant of Sarathi-serve that uses shortest prefill first ordering, referred to as Sarathi-serve (SPF).

Similarly, we assess SLAI under both FCFS and SPF prefill orderings, referred as SLAI (FCFS, fixed offset) and SLAI (SPF, fixed offset), respectively. In addition, we evaluate a dynamic version of SLAI, called SLAI (SPF, dynamic offset), where the offset $\delta$ is adjusted at runtime based on GPU memory utilization measurements. When GPU memory usage is low, a small offset is used to allow more prefill-phase requests into the system. When memory usage is high, a larger offset is applied so that decode-iterations are marked critical earlier and prioritized accordingly, thus clearing out memory. For completeness, we also include vLLM in our comparisons, a scheduler that prioritizes prefill-phase requests and serves as another relevant baseline.

***Metrics.*** We evaluate two key metrics. The first measures the median TTFT since this is measured only once per request. This reflects how well the scheduler meets responsiveness objectives across requests. The second metric is the 99th percentile of the TBT, which is computed once per generated token and captures tail latency during the decode-iterations. This helps assess how smoothly tokens are generated over time.

***Workload.*** To emulate realistic traffic, we generate synthetic traces based on request length distributions observed in the `openchat_shareGPT4` dataset (see Table 1). Each synthetic request is generated by sampling one row from the dataset, which provides the number of prefill tokens and decode tokens for that request. We then cap the total length at 8192 tokens to respect our maximum sequence length. The sampled prefill and decode lengths are used directly in our trace generator. Inference requests are generated according to a Poisson process with rates $\lambda \in \{0.2, 0.4, \dots, 1.6\}$ requests per second. Each experiment runs for 35 minutes of wall-clock time, including a short warm-up period before measurement begins. To ensure reproducibility, we fix random seeds for both the trace generation and scheduler runs. We consider two types of requests associated with paying and free-tier users. Paying users expect faster and smoother generation compared to free-tier users. To reflect this, we assign a TBT SLO threshold of 0.1 seconds for paying users and 0.5 seconds for free-tier user. These values are slightly relaxed compared to real-world production settings because our implementation is in Python (which is not fully optimized), includes telemetry overhead, and also reflects the inherent performance limitations of the model–hardware combination. Each incoming request is randomly marked as associated with a paying or free-tier user with some probability.

| **Dataset** | **Prompt length (tokens)** | | | **Decode length (tokens)** | | |
|---|---|---|---|---|---|---|
| | **Median** | **P90** | **Std.** | **Median** | **P90** | **Std.** |
| openchat_sharegpt4 | 1730 | 5696 | 2088 | 415 | 834 | 101 |

Table 1. Prompt and decode length (token) statistics for requests in the `openchat_sharegpt4` dataset.

***Results discussion.*** We first consider a scenario where each request has a 5% chance of coming from a paying user. This low percentage reflects the user distribution seen on platforms like ChatGPT, where most users belong to the free tier and Figures 7a and 7b present the 99th percentile

TBT (P99 TBT) for paying and free-tier users, respectively. Figure 7c illustrates the median TTFT as a function of the request rate. In all experiments, we configure Sarathi-serve (both FCFS and SPF variants) with a token budget of 512 to ensure that the 0.1-second TBT target for paying users is met. For all SLAI variants, we use the same token budget, and set both the number of active requests and concurrent decode-phase request limit to 128. For SLAI (FCFS/SPF, fixed offset), we set the offset parameter to 10, which controls when a decode-phase request becomes time-critical, whereas for SLAI (SPF, dynamic offset), the offset is set to 5 if GPU memory utilization is below 96%, and to 10 otherwise.

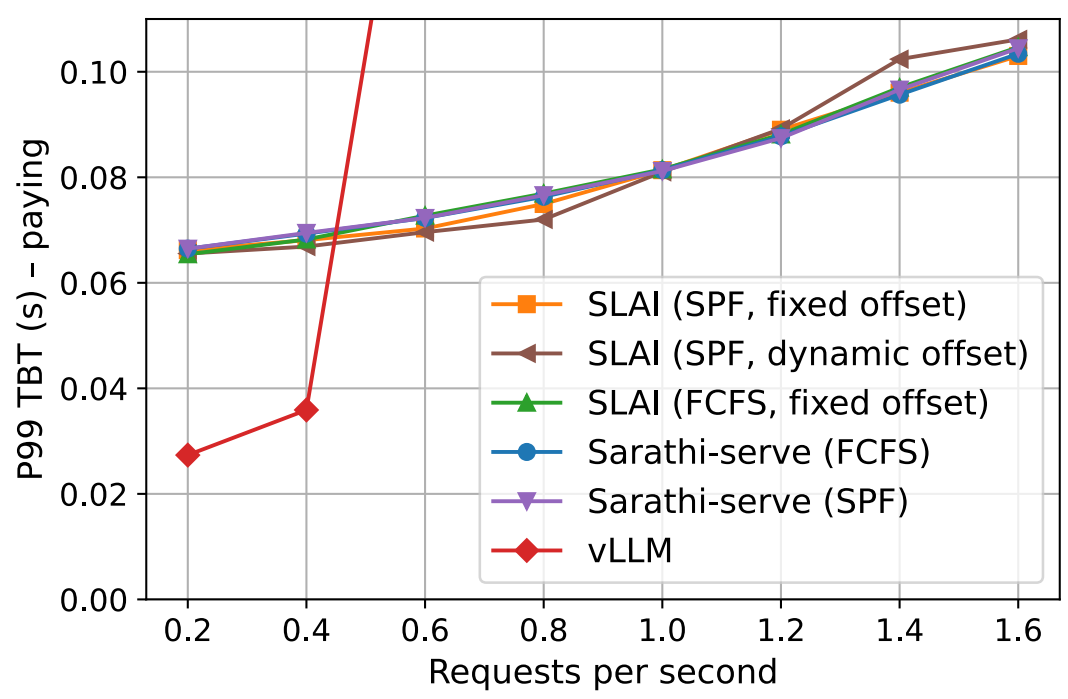

(a) 99th percentile TBT for paying users across different request rates for a target of 0.1 seconds.

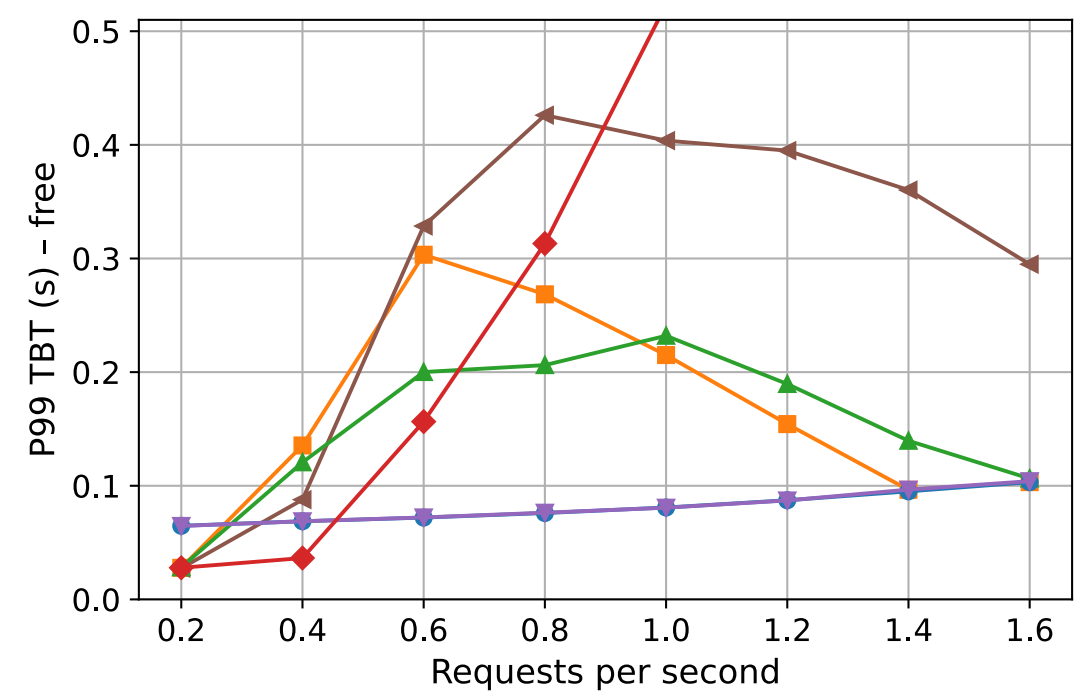

(b) 99th percentile TBT for free-tier users across different request rates for a target of 0.5 seconds.

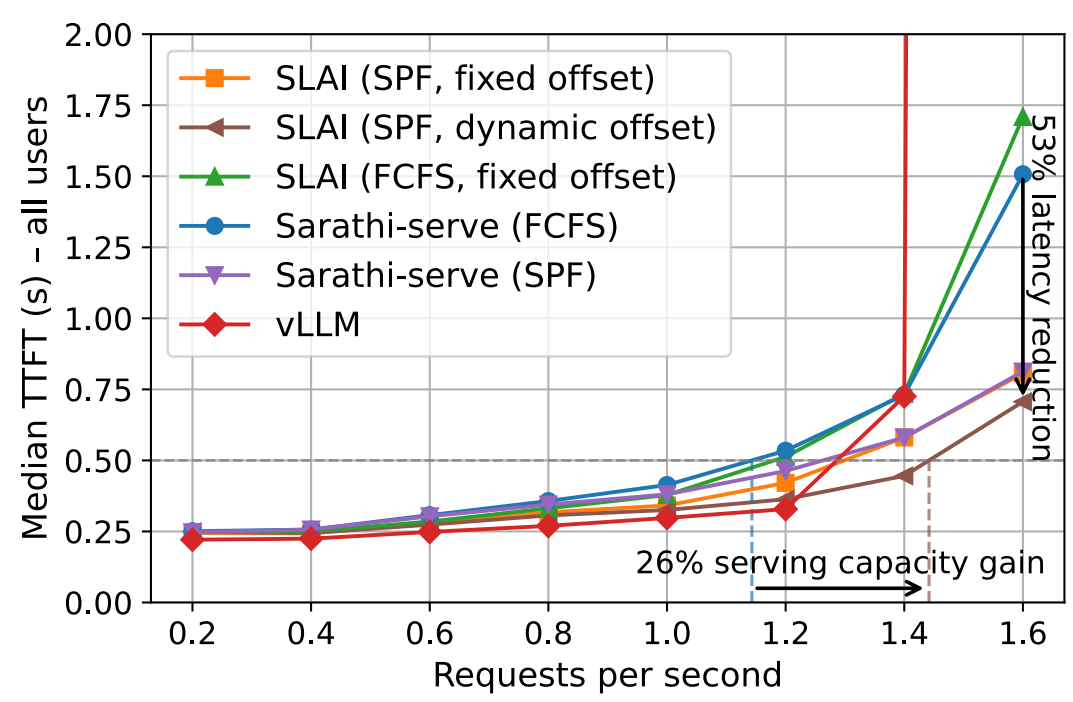

(c) Median TTFT for all users as a function of request rate.

Fig. 7. Performance comparison of SLAI, Sarathi-serve, and vLLM under mixed user workloads with 5% paying users. SLAI (SPF, dynamic offset) achieves the best latency-throughput trade-off.

***TBT Behavior.*** Figures 7a and 7b show how SLAI dynamically prioritizes requests during their decode phase based on their TBT targets. Under Sarathi-serve, the 99th percentile TBT steadily increases for both paying and free-tier requests as the system load grows. This happens because every decode-iteration is included in every batch, and as load increases, so does the batch execution time, leading to higher delays for all requests. By contrast, SLAI handles decode-iterations differently. Requests from paying users have strict (low) TBT targets, which in most cases are always considered time-critical. As the load increases and batches take longer to run, their P99 TBT naturally increases.

Free-tier requests, however, have more relaxed TBT targets. At lower loads, the scheduler can defer these decode-iterations to prioritize prefills, since they are not immediately time-critical. This initially causes their P99 TBT to rise. But as the load continues to increase and batch execution time becomes longer, the window during which a free-tier decode-iteration remains non-critical shortens. As a result, these decode-iterations become time-critical sooner and are prioritized earlier in scheduling. This leads to a drop in their P99 TBT. Eventually, at high loads, all free-tier decode-iterations are immediately marked as time-critical, and their TBT increases again—now dominated by the growing batch execution time, similar to paying users. Lastly, vLLM since it is a prefill-prioritizing scheduler ends up violating P99 TBT at a relatively low load and thus is not effective at managing decode-phase requests.

***TTFT behaviour.*** Figure 7c shows the median TTFT as a function of requests per second. The vLLM policy, which prioritizes prefill-phase requests, achieves the lowest median TTFT at low loads. However, it does so by aggressively batching prefill requests at the expense of violating 99th percentile TBT latency constraints, making it unsuitable for scenarios with strict QoS requirements. Sarathi-serve improves upon this by balancing prefill and decode phases to maintain both acceptable median TTFT and TBT tail latencies. When Sarathi-serve is combined with the SPF-based policy it yields a better median TTFT than its FCFS counterpart, highlighting the benefit of reordering prefill requests by prompt length. However, Sarathi-serve does not adapt to heterogeneous TBT deadlines across requests. SLAI (SPF, fixed offset) addresses this by selectively deferring decode-phase requests with relaxed deadlines, achieving further improvements in median TTFT. Finally, SLAI (SPF, dynamic offset) introduces dynamic decode-iteration deferral based on real-time GPU memory utilization, allowing the system to better utilize available token budget of a batch. As a result, SLAI delivers significant performance improvements: it reduces the median TTFT from 1.5 seconds (under Sarathi-Serve (FCFS)) to 0.7 seconds—a 53% improvement under high load—and increases the maximum sustainable request rate from 1.15 to 1.45 requests per second while maintaining a fixed median TTFT constraint of 0.5 seconds and meeting tail TBT latency targets, representing a 26% increase in serving capacity.

See Appendix F for additional experimental results that highlight several important aspects: i) the performance of different policies as a function of prompt lengths, ii) the impact of prioritizing paying users over free-tier users during the prefill phase, and iii) how the policies compare when the proportion of paying users increases to 50% or 95%.

### 7.1 Choosing the parameters for SLAI (SPF, dynamic offset)

***Role of*** $\delta$***.*** As described in Section 6, the last schedulable time for the $i^{\text{th}}$ decode-iteration of request $j$ is given by:

$$C_{i,j} = e_{i-1,j} + \text{TBT}_j - \delta \cdot \bar{t}_{\text{batch}} \tag{9}$$

At the time of constructing batch $m$ (at time $t_m$), we mark a decode-iteration as *time-critical* if $t_m \geq C_{i,j}$. Otherwise, it is considered *non-critical* and can be deferred to a future batch. The offset parameter $\delta$ controls how aggressively we prioritize decode-iterations of decode-phase requests: a smaller $\delta$ delays when decode-iterations are considered critical (favoring prefill-phase requests), while a larger $\delta$ marks them critical sooner (favoring decode-phase requests).

***Empirical motivation.*** Recall that our goal is to reduce the median TTFT while ensuring that tail TBT constraints are always respected. Figure 8 shows GPU memory utilization under three different policies: SLAI (SPF, fixed offset 10), SLAI (SPF, fixed offset 5), and SLAI (SPF, dynamic offset). All experiments were conducted at a load of 1.6 requests per second.

With a fixed offset of $\delta = 10$, decode-phase iterations are prioritized. This keeps TBT well within bounds but underutilizes the GPU memory. For example, requests from users with a relaxed TBT constraint of 0.5 seconds consistently achieve tail TBT values far below the target (see Fig. 7b). This suggests we are being too conservative and missing the chance to initiate more prefill-phase requests.

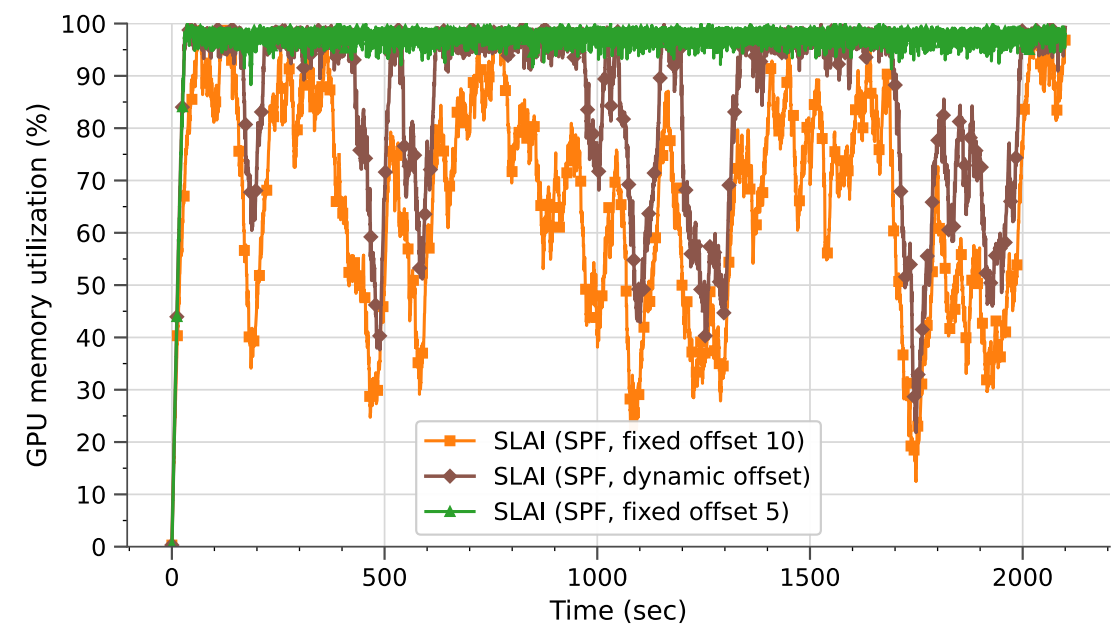

Fig. 8. GPU memory utilization over time for SLAI (SPF, fixed offset 10), SLAI (SPF, fixed offset 5), and SLAI (SPF, dynamic offset: $\delta = 5$ below 96%, $\delta = 10$ otherwise) for a fixed load of 1.6 requests per second.

We then tried a fixed offset of $\delta = 5$, which gives higher priority to prefill-phase requests and delays scheduling of decode-iterations of decode-phase requests. However, at high load (e.g., 1.6 requests per second), this led to memory saturation, as seen in Fig. 8. This happens because decode-phase requests' KV tensors stayed in memory for too long, preventing the system from admitting new prefill-phase requests. As a result, TTFT blows up, and the SLAI policy with fixed offset 5 is unable to sustain the 1.6 requests per second load.

These results show a clear trade-off: we want to be aggressive in admitting prefill-phase requests when memory allows, but we also need to ensure decode-phase requests do not build up and block future admissions.

***Dynamic offset policy.*** To balance these needs, we propose a simple dynamic policy that switches $\delta$ based on current GPU memory usage:

$$\delta = \begin{cases} 5, & \text{if GPU memory utilization } < 96\% \quad \text{(favor prefill-phase requests);} \\ 10, & \text{otherwise} \quad \text{(favor decode-phase iterations).} \end{cases}$$

When memory usage is below 96%, we assume there is enough headroom to safely admit more prefill-phase requests. When usage crosses 96%, we shift priority to decode-phase iterations to eventually complete requests and free up memory.

The 96% threshold was chosen based on empirical observations. In our setup, each new prefill-phase request is allocated memory for its KV tensors at the time of admission—even before all prefill tokens are processed. We found that admitting a new prefill-phase request can consume up to 4% of GPU memory. To avoid admission failures, we ensure that this much headroom is always available, and thus set the switching threshold to 96%.

***Observed effect.*** As shown in Figure 8, the dynamic policy maintains high GPU memory utilization similar to fixed $\delta = 5$, but avoids hitting 100%, which can cause admission stalls. This dynamic adjustment ensures that TBT constraints are met while also improving TTFT and overall system throughput.

## 8 Conclusion

This paper presented a framework for designing efficient LLM inference systems. By modeling the unique two-phase structure of LLM inference, we identified two key design principles—optimal tiling and optimal resource allocation—as essential for achieving high throughput. Based on these insights, we introduced the RAD scheduler, which, when combined with a uniformly random resource planner, provably achieves throughput optimality. In real-world deployments, however, systems must also meet SLOs. We argued that practical schedulers often approximate only one of the two design principles, thereby sacrificing some throughput to satisfy these SLOs. To handle heterogeneous request classes with different latency needs, we proposed the SLAI scheduler. SLAI reduces TTFT by intelligently prioritizing requests while still meeting tail TBT constraints. In comparison to existing state-of-the-art, SLAI reduced the median TTFT by 53% and increased the maximum serving capacity by 26% for a fixed median TTFT, while meeting the TBT constraints.

## Acknowledgments

This work was supported in part by NSF Award CNS-2212202 and NSF Award 2148224, which receives funding from OUSD R&E, NIST, and industry partners through the Resilient & Intelligent NextG Systems (RINGS) program. We would also like to thank the reviewers and shepherd of our ACM SIGMETRICS 2026 submission. Their thoughtful feedback and suggestions were instrumental in improving the quality and clarity of this paper.

## References

[1] 2023. *Introducing AMD CDNA™ 3 Architecture*. White Paper. Advanced Micro Devices, Inc. https://www.amd.com/content/dam/amd/en/documents/instinct-tech-docs/white-papers/amd-cdna-3-white-paper.pdf Accessed: 2025-06-18.

[2] Amey Agrawal, Nitin Kedia, Ashish Panwar, Jayashree Mohan, Nipun Kwatra, Bhargav Gulavani, Alexey Tumanov, and Ramachandran Ramjee. 2024. Taming Throughput-Latency Tradeoff in LLM Inference with Sarathi-Serve. In *18th USENIX Symposium on Operating Systems Design and Implementation (OSDI 24)*. USENIX Association, Santa Clara, CA, 117–134. https://www.usenix.org/conference/osdi24/presentation/agrawal

[3] Joshua Ainslie, James Lee-Thorp, Michiel De Jong, Yury Zemlyanskiy, Federico Lebrón, and Sumit Sanghai. 2023. Gqa: Training generalized multi-query transformer models from multi-head checkpoints. *arXiv preprint arXiv:2305.13245* (2023).

[4] NVIDIA Corporation. 2025. NVIDIA Dynamo. https://developer.nvidia.com/dynamo. Accessed: 2025-07-07.

[5] Tri Dao, Daniel Y. Fu, Stefano Ermon, Atri Rudra, and Christopher Ré. 2022. FlashAttention: Fast and Memory-Efficient Exact Attention with IO-Awareness. arXiv:2205.14135 [cs.LG] https://arxiv.org/abs/2205.14135

[6] Frederic G Foster. 1953. On the stochastic matrices associated with certain queuing processes. *The Annals of Mathematical Statistics* 24, 3 (1953), 355–360.

[7] Connor Holmes, Masahiro Tanaka, Michael Wyatt, Ammar Ahmad Awan, Jeff Rasley, Samyam Rajbhandari, Reza Yazdani Aminabadi, Heyang Qin, Arash Bakhtiari, Lev Kurilenko, and Yuxiong He. 2024. DeepSpeed-FastGen: High-throughput Text Generation for LLMs via MII and DeepSpeed-Inference. arXiv:2401.08671 [cs.PF] https://arxiv.org/abs/2401.08671

[8] Ke Hong, Guohao Dai, Jiaming Xu, Qiuli Mao, Xiuhong Li, Jun Liu, Kangdi Chen, Yuhan Dong, and Yu Wang. 2024. FlashDecoding++: Faster Large Language Model Inference on GPUs. arXiv:2311.01282 [cs.LG] https://arxiv.org/abs/2311.01282

[9] Yanping Huang, Youlong Cheng, Ankur Bapna, Orhan Firat, Mia Xu Chen, Dehao Chen, HyoukJoong Lee, Jiquan Ngiam, Quoc V. Le, Yonghui Wu, and Zhifeng Chen. 2019. *GPipe: efficient training of giant neural networks using pipeline parallelism*. Curran Associates Inc., Red Hook, NY, USA.

[10] Intel Corporation. 2024. *Intel® Gaudi® 3 AI Accelerator White Paper*. Technical Report. Intel Corporation. https://www.intel.com/content/www/us/en/content-details/817486/intel-gaudi-3-ai-accelerator-white-paper.html Accessed: 2025-06-18.

[11] Kunal Jain, Anjaly Parayil, Ankur Mallick, Esha Choukse, Xiaoting Qin, Jue Zhang, Íñigo Goiri, Rujia Wang, Chetan Bansal, Victor Rühle, Anoop Kulkarni, Steve Kofsky, and Saravan Rajmohan. 2025. Intelligent Router for LLM Workloads: Improving Performance Through Workload-Aware Load Balancing. arXiv:2408.13510 [cs.DC] https://arxiv.org/abs/2408.13510

[12] Albert Q. Jiang, Alexandre Sablayrolles, Arthur Mensch, Chris Bamford, Devendra Singh Chaplot, Diego de las Casas, Florian Bressand, Gianna Lengyel, Guillaume Lample, Lucile Saulnier, Lélio Renard Lavaud, Marie-Anne Lachaux, Pierre Stock, Teven Le Scao, Thibaut Lavril, Thomas Wang, Timothée Lacroix, and William El Sayed. 2023. Mistral 7B. arXiv:2310.06825 [cs.CL] https://arxiv.org/abs/2310.06825

[13] Jordan Juravsky, Bradley Brown, Ryan Ehrlich, Daniel Y. Fu, Christopher Ré, and Azalia Mirhoseini. 2024. Hydragen: High-Throughput LLM Inference with Shared Prefixes. arXiv:2402.05099 [cs.LG] https://arxiv.org/abs/2402.05099

[14] Aditya K. Kamath, Ramya Prabhu, Jayashree Mohan, Simon Peter, Ramachandran Ramjee, and Ashish Panwar. 2025. POD-Attention: Unlocking Full Prefill-Decode Overlap for Faster LLM Inference. In *Proceedings of the 30th ACM International Conference on Architectural Support for Programming Languages and Operating Systems, Volume 2 (ASPLOS '25)*. ACM, 897–912. doi:10.1145/3676641.3715996

[15] Woosuk Kwon, Zhuohan Li, Siyuan Zhuang, Ying Sheng, Lianmin Zheng, Cody Hao Yu, Joseph E. Gonzalez, Hao Zhang, and Ion Stoica. 2023. Efficient Memory Management for Large Language Model Serving with PagedAttention. arXiv:2309.06180 [cs.LG] https://arxiv.org/abs/2309.06180

[16] Yaniv Leviathan, Matan Kalman, and Yossi Matias. 2023. Fast Inference from Transformers via Speculative Decoding. arXiv:2211.17192 [cs.LG] https://arxiv.org/abs/2211.17192

[17] Yueying Li, Jim Dai, and Tianyi Peng. 2025. Throughput-optimal scheduling algorithms for llm inference and ai agents. *arXiv preprint arXiv:2504.07347* (2025).

[18] Yuhan Liu, Hanchen Li, Yihua Cheng, Siddhant Ray, Yuyang Huang, Qizheng Zhang, Kuntai Du, Jiayi Yao, Shan Lu, Ganesh Ananthanarayanan, Michael Maire, Henry Hoffmann, Ari Holtzman, and Junchen Jiang. 2024. CacheGen: KV Cache Compression and Streaming for Fast Large Language Model Serving. arXiv:2310.07240 [cs.NI] https://arxiv.org/abs/2310.07240

[19] Michael Mitzenmacher and Rana Shahout. 2025. Queueing, Predictions, and LLMs: Challenges and Open Problems. *arXiv preprint arXiv:2503.07545* (March 2025). arXiv:2503.07545 [cs.AI] https://arxiv.org/abs/2503.07545

[20] Moonmoon Mohanty, Gautham Bolar, Preetam Patil, UmaMaheswari Devi, Felix George, Pratibha Moogi, and Parimal Parag. 2025. Deferred prefill for throughput maximization in LLM inference. In *Proceedings of The 5th Workshop on Machine Learning and Systems (EuroMLSys'2025)*. ACM, New York, NY, USA, 100–106. doi:10.1145/3721146.3721962

[21] Deepak Narayanan, Mohammad Shoeybi, Jared Casper, Patrick LeGresley, Mostofa Patwary, Vijay Korthikanti, Dmitri Vainbrand, Prethvi Kashinkunti, Julie Bernauer, Bryan Catanzaro, Amar Phanishayee, and Matei Zaharia. 2021. Efficient large-scale language model training on GPU clusters using megatron-LM. In *Proceedings of the International Conference for High Performance Computing, Networking, Storage and Analysis* (St. Louis, Missouri) *(SC '21)*. Association for Computing Machinery, New York, NY, USA, Article 58, 15 pages. doi:10.1145/3458817.3476209

[22] NVIDIA. [n. d.]. Matrix Multiplication Background User's Guide. https://docs.nvidia.com/deeplearning/performance/dl-performance-matrix-multiplication/index.html. Accessed: February 27, 2025.

[23] NVIDIA. 2025. FasterTransformer. https://github.com/NVIDIA/FasterTransformer

[24] NVIDIA. 2025. TensorRT-LLM: A TensorRT toolbox for optimized large-language-model inference. https://github.com/NVIDIA/TensorRT-LLM

[25] NVIDIA Corporation. 2020. *NVIDIA A100 Tensor Core GPU Architecture*. Technical Report. NVIDIA Corporation. https://images.nvidia.com/aem-dam/en-zz/Solutions/data-center/nvidia-ampere-architecture-whitepaper.pdf Version 1.0.

[26] OpenAI. 2025. ChatGPT. https://chat.openai.com

[27] Noam Shazeer. 2019. Fast transformer decoding: One write-head is all you need. *arXiv preprint arXiv:1911.02150* (2019).

[28] Karl Sigman and Ronald W Wolff. 1993. A review of regenerative processes. *SIAM review* 35, 2 (1993), 269–288.

[29] Biao Sun, Ziming Huang, Hanyu Zhao, Wencong Xiao, Xinyi Zhang, Yong Li, and Wei Lin. 2024. Llumnix: Dynamic Scheduling for Large Language Model Serving. In *18th USENIX Symposium on Operating Systems Design and Implementation (OSDI 24)*. USENIX Association, Santa Clara, CA, 173–191. https://www.usenix.org/conference/osdi24/presentation/sun-biao

[30] Ashish Vaswani, Noam Shazeer, Niki Parmar, Jakob Uszkoreit, Llion Jones, Aidan N Gomez, Łukasz Kaiser, and Illia Polosukhin. 2017. Attention is all you need. *Advances in neural information processing systems* 30 (2017).

[31] Guan Wang, Sijie Cheng, Xianyuan Zhan, Xiangang Li, Sen Song, and Yang Liu. 2024. OpenChat: Advancing Open-source Language Models with Mixed-Quality Data. arXiv:2309.11235 [cs.CL] https://arxiv.org/abs/2309.11235

[32] Zihao Ye, Lequn Chen, Ruihang Lai, Wuwei Lin, Yineng Zhang, Stephanie Wang, Tianqi Chen, Baris Kasikci, Vinod Grover, Arvind Krishnamurthy, and Luis Ceze. 2025. FlashInfer: Efficient and Customizable Attention Engine for LLM Inference Serving. arXiv:2501.01005 [cs.DC] https://arxiv.org/abs/2501.01005

[33] Gyeong-In Yu, Joo Seong Jeong, Geon-Woo Kim, Soojeong Kim, and Byung-Gon Chun. 2022. Orca: A distributed serving system for {Transformer-Based} generative models. In *16th USENIX Symposium on Operating Systems Design and Implementation (OSDI 22)*. 521–538.

[34] Yilong Zhao, Shuo Yang, Kan Zhu, Lianmin Zheng, Baris Kasikci, Yang Zhou, Jiarong Xing, and Ion Stoica. 2024. BlendServe: Optimizing Offline Inference for Auto-regressive Large Models with Resource-aware Batching. *arXiv preprint arXiv:2411.16102* (2024).

[35] Lianmin Zheng, Liangsheng Yin, Zhiqiang Xie, Chuyue Sun, Jeff Huang, Cody Hao Yu, Shiyi Cao, Christos Kozyrakis, Ion Stoica, Joseph E. Gonzalez, Clark Barrett, and Ying Sheng. 2024. SGLang: Efficient Execution of Structured Language Model Programs. arXiv:2312.07104 [cs.AI] https://arxiv.org/abs/2312.07104

[36] Yinmin Zhong, Shengyu Liu, Junda Chen, Jianbo Hu, Yibo Zhu, Xuanzhe Liu, Xin Jin, and Hao Zhang. 2024. {DistServe}: Disaggregating prefill and decoding for goodput-optimized large language model serving. In *18th USENIX Symposium on Operating Systems Design and Implementation (OSDI 24)*. 193–210.

# A Other Related Work

## A.1 Cluster-level Routing

Per-GPU schedulers rely on the router to provide a well-balanced stream of requests. Most production systems still use simple strategies like round-robin or shortest-queue routing. These methods overlook the complex relationship between request length, prompt size, and current GPU state. The Intelligent Router for LLM Workloads [11] addresses this by framing routing as a sequential decision problem. It trains a workload-aware reinforcement learning agent to minimize overall latency by predicting each request's response length and estimating how much delay each placement would cause.

## A.2 KV-Cache Management

In transformer models, self-attention reuses all previous tokens, making KV-cache management critical for both speed and capacity.

***Memory layout.*** vLLM [15] uses paged attention, which divides GPU memory into fixed-size blocks that are dynamically assigned to requests. This reduces fragmentation and allows hundreds of requests to run in parallel. Llumnix [29] builds on this by migrating KV tensors across replicas in real time, balancing memory usage and lowering preemption costs.

***Prefix reuse and compression.*** Some systems try to reduce the amount of KV data stored or recomputed. SGLang [35] introduces a radix-tree cache and orders batches to maximize prefix reuse across multi-turn chats and speculative decoding. CacheGen [18] compresses KV blocks and streams them on demand, while FlashInfer [32] creates custom GPU kernels that operate directly on the compressed format. These techniques are independent of scheduling and routing and can be used alongside paged layouts or distributed setups.

***Preemption strategies.*** When GPU memory runs out, systems must either recompute or offload paused requests. vLLM [15] and Sarathi-Serve [2] evict stalled requests, splice their outputs back into the prompt, and later rebuild the KV cache. DistServe [36], on the other hand, moves the KV tensors to host memory and resumes decoding once space is available. Each method involves a trade-off between memory traffic and computation, and interacts closely with the scheduler's design.

## A.3 Speculative decoding

Speculative decoding [16] offers a complementary approach that accelerates decode-phase computation by generating token drafts using a smaller auxiliary model, which are then validated by the larger target model. While originally proposed to reduce per-request latency, this technique can also benefit scheduling by reducing decode durations, improving GPU throughput, and enabling more efficient batch formation under tight latency constraints.

# B Upper Bound Proof

In Section B.1, we establish a lower bound on the time required for any LLM inference system to serve a given set of requests. Building on this, Section B.2 derives a lower bound on the number of pending requests at a given time in any system. Finally, in Section B.3, we use these results to prove Theorem 1, showing that if the request arrival rate exceeds $r/\bar{T}^R$, the number of pending requests grows linearly over time—thereby establishing an upper bound on the system's capacity region..

### B.1 Request Set Drain Time

The first part of the proof is to consider a more powerful class of resource planners and schedulers (as described below) and lower bound the time required by them to start and complete $k$ requests.

Consider a set, $\mathcal{R} = \{R_1, \ldots, R_k\}$, of $k$ requests, where each request $R_j$'s prompt and output lengths are, $L_j^P, L_j^D \sim p_{L^P,L^D}$, and are i.i.d., across requests. Further, consider a more powerful class of resource planners and schedulers which are able to, a) commence serving of all requests at time 0, b) serve prefill and decode iterations of requests in an arbitrary order, c) serve the different iterations of requests at any node, even simultaneously, while paying zero cost for request transfers. Clearly, the realistic class of schedulers considered in Theorem 1 which— a) may commence serving requests once they arrive, b) have to serve prefill and decode iterations in order according to Definitions 1 and 2, and, c) may only transfer requests once the respective computations at the source node are completed— are a subset of this more powerful class of systems. To develop our bounds, we will compute the least time required by this powerful class of systems to completely serve all the $k$ requests.

Consider a sequence of batches $\left(\mathcal{B}_{m,r'}\right)_{m=1}^{M_{r'}}$ at node $r'$, for every $r' \in \{1, \ldots, r\}$. Let the set of batch sequences completely serve all the $k$ requests. Then, the request set drain time which is the time required to complete $k$ requests using all $r$ nodes, denoted $T(\mathcal{R})$, is given by,

$$\begin{aligned} T(R) &= \max_{r' \in \{1,\ldots,r\}} \sum_{m=1}^{M_{r'}} T^B(\mathcal{B}_{m,r'}), \\ &\geq \frac{1}{r} \sum_{r'=1}^{r} \sum_{m=1}^{M_{r'}} T^B(\mathcal{B}_{m,r'}). \end{aligned} \tag{10}$$

Let $\tau(\mathcal{B})$ be the token count of batch $\mathcal{B}$. Let $\mathcal{S}^P(\mathcal{B})$ and $\mathcal{S}^D(\mathcal{B})$ be the sets of prefill and decode phase requests in the batch. For a decode-phase request $R_j$ let $i_j$ denote the token position of its decode-iteration in the batch. And, similarly for a prefill-phase request $R_j$, let $i_j$ and $c_j$ denote the starting token index and chunk size of its prefill-iteration in the batch. Let $t_{\text{row}}$, $t_{\text{col}}$ denote the output tile dimensions and $t_{\text{red}}$ denotes the reduction tile dimension for the GeMM. Recall the batch computation time formula for a batch $\mathcal{B}$,

$$\begin{aligned} T^B(\mathcal{B}) = &\frac{1}{\mu_{\text{Lin}}(t_{\text{row}}, t_{\text{col}}, t_{\text{red}})} \left\lceil \frac{\tau(\mathcal{B})}{t_{\text{col}}} \right\rceil + \frac{1}{\mu_{\text{nLin}}} \tau(\mathcal{B}) + N \sum_{j:R_j \in \mathcal{S}^D(\mathcal{B})} T^{D,SA}(i_j) \\ &+ \frac{N}{s\mu(t_{\text{row}}, t_{\text{col}}, t_{\text{red}})} \sum_{j:R_j \in \mathcal{S}^P(\mathcal{B})} \left( \left\lceil \frac{i_j + c_j - 1}{t_{\text{row}}} \right\rceil \left\lceil \frac{c_j}{t_{\text{col}}} \right\rceil \frac{d}{t_{\text{red}}} + \frac{d}{t_{\text{row}}} \left\lceil \frac{c_j}{t_{\text{col}}} \right\rceil \left\lceil \frac{i_j + c_j - 1}{t_{\text{red}}} \right\rceil \right), \end{aligned}$$

Then,

$$\begin{aligned}T_B(\mathcal{B}) &\overset{(a)}{\geq} \frac{1}{\mu_{\text{Lin}}(t_{\text{row}}, t_{\text{col}}, t_{\text{red}})}\frac{\tau(\mathcal{B})}{t_{\text{col}}} + \frac{1}{\mu_{\text{nLin}}}\tau(\mathcal{B}) + N\sum_{j:R_j\in\mathcal{S}^D(\mathcal{B})} T^{D,SA}(i_j)\\ &+ \frac{N}{s\mu(t_{\text{row}}, t_{\text{col}}, t_{\text{red}})}\sum_{j:R_j\in\mathcal{S}^P(\mathcal{B})}\left(\frac{i_j+c_j-1}{t_{\text{row}}}\frac{c_j}{t_{\text{col}}}\frac{d}{t_{\text{red}}} + \frac{d}{t_{\text{row}}}\frac{c_j}{t_{\text{col}}}\frac{i_j+c_j-1}{t_{\text{red}}}\right),\\ &\overset{(b)}{\geq} \frac{1}{\mu_{\text{Lin}}(t^*_{\text{row}}, t^*_{\text{col}}, t^*_{\text{red}})}\frac{\tau(\mathcal{B})}{t^*_{\text{col}}} + \frac{1}{\mu_{\text{nLin}}}\tau(\mathcal{B}) + N\sum_{j:R_j\in\mathcal{S}^D(\mathcal{B})} T^{D,SA}(i_j)\\ &+ \frac{N}{s\mu(t^*_{\text{row}}, t^*_{\text{col}}, t^*_{\text{red}})}\sum_{j:R_j\in\mathcal{S}^P(\mathcal{B})}\left(\frac{i_j+c_j-1}{t^*_{\text{row}}}\frac{c_j}{t^*_{\text{col}}}\frac{d}{t^*_{\text{red}}} + \frac{d}{t^*_{\text{row}}}\frac{c_j}{t^*_{\text{col}}}\frac{i_j+c_j-1}{t^*_{\text{red}}}\right),\end{aligned} \tag{11}$$

where (a) follows by simply removing the ceiling function, and (b) follows from Assumption 1.

Consider request $R_j$. Let $t^*_{\text{lcm}} = \text{LCM}(t^*_{\text{row}}, t^*_{\text{col}}, t^*_{\text{red}})$. By Assumption 3, $L^P_j$ is a multiple of $t^*_{\text{lcm}}$. Therefore, the optimal set of chunk positions and chunk size to complete the self-attention in the prefill phase in minimal time is,

$$\mathcal{I}_j = \left\{(1, t^*_{\text{lcm}}, (t^*_{\text{lcm}}+1, t^*_{\text{lcm}}), \ldots, (L_P - t^*_{\text{lcm}}, t^*_{\text{lcm}})\right\}.$$

Then, define its "optimal effective completion time" as,

$$\begin{aligned}T^R_*(j) &= \frac{1}{\mu_{\text{Lin}}(t^*_{\text{row}}, t^*_{\text{col}}, t^*_{\text{red}})}\frac{L^P_j + L^D_j}{t^*_{\text{col}}} + \frac{1}{\mu_{\text{nLin}}}(L^P_j + L^D_j) + N\sum_{i=L^P_j+1}^{L^P_j+L^D_j} T^{D,SA}(i)\\ &+ \frac{N}{s\mu(t^*_{\text{row}}, t^*_{\text{col}}, t^*_{\text{red}})}\sum_{(i_j,c_j)\in\mathcal{I}_j}\left(\frac{i_j+c_j-1}{t^*_{\text{row}}}\frac{c_j}{t^*_{\text{col}}}\frac{d}{t^*_{\text{red}}} + \frac{d}{t^*_{\text{row}}}\frac{c_j}{t^*_{\text{col}}}\frac{i_j+c_j-1}{t^*_{\text{red}}}\right),\end{aligned} \tag{12}$$

$$\begin{aligned}&= \frac{1}{\mu_{\text{Lin}}(t^*_{\text{row}}, t^*_{\text{col}}, t^*_{\text{red}})}\frac{L^P_j + L^D_j}{t^*_{\text{col}}} + \frac{1}{\mu_{\text{nLin}}}(L^P_j + L^D_j) + N\sum_{i=L^P_j+1}^{L^P_j+L^D_j} T^{D,SA}(i)\\ &+ \frac{Nd}{s\mu(t^*_{\text{row}}, t^*_{\text{col}}, t^*_{\text{red}})t^*_{\text{row}}t^*_{\text{col}}t^*_{\text{red}}}L^P_j\left(L^P_j + t^*_{\text{lcm}}\right).\end{aligned} \tag{13}$$

Then, we have the following alternate lower bound for the request set drain time.

Proposition 1. $T(\mathcal{R}) \geq \frac{1}{r}\sum_{j=1}^k T^R_*(j)$.

Proof. From (10) we have,

$$T(\mathcal{R}) \geq \frac{1}{r}\sum_{r'=1}^{r}\sum_{m=1}^{M_{r'}} T^B\left(\mathcal{B}_{m,r'}\right),$$

where recall that $\left(\mathcal{B}_{m,r'} : r' \in \{1,\ldots,r\}, m \in \{1,\ldots,M_{r'}\}\right)$ is a set of batches that complete all the prefill and decode iterations of all the $k$ requests in $\mathcal{R}$. The batch execution time of a batch is lower-bounded in (11). By rearranging the terms in the summation of batch execution times, and using (12), we obtain the result. □

Then, the following limit follows by the strong law of large numbers.

Proposition 2. *For any $k \in \mathbb{N}$, let the set of $k$ requests $R_1, \ldots, R_k$ have prompt and output lengths distributed according to $p_{L^P,L^D}$, and let these lengths be i.i.d., across requests. Then,*

$$\lim_{k\to\infty} \frac{\frac{1}{r}\sum_{j=1}^{k} T_*^R(j)}{k} = \frac{\bar{T}^R}{r},$$

*where $\bar{T}^R$ is as in Theorem 1.*

### B.2 A lower bound on the pending requests in the system

Let $\left(\mathcal{A}^{\text{cont.}}(t)\right)_{t\geq 0}$ denote the arrival process of requests to the system, where $\mathcal{A}^{\text{cont.}}(t)$ denotes the set of requests that have arrived by time $t$. Let, $A(t) \triangleq \left|\mathcal{A}^{\text{cont.}}(t)\right|$ denote the number of arrivals by time $t$. Then, recall that by the premise of the theorem, $(A(t))_{t\geq 0}$ is a Poisson point process of rate $\lambda$, and each request has prompt and output lengths distributed according to $p_{L^P,L^D}$ i.i.d., across requests. Finally, recall that $Q(t)$ denotes the number of pending requests in the system at time $t$.

Consider the worst-case amount of time, $T^{\max}$, required to complete a request. This is when the prompt and output lengths are the longest possible, i.e., $l^{P,\max}$ and $l^{D,\max}$. Moreover, in this worst case, the scheduler schedules prefill-iterations with chunk size 1 at a time without batching to complete the prefill-phase, and schedules all the decode-iterations as well without batching. Therefore, we have,

$$T^{\max} = \max_{(t_{\text{row}},t_{\text{col}})\in\mathcal{T}_{\text{out}},\, t_{\text{red}}\in\mathcal{T}_{\text{red}}} \left[ \frac{l^{P,\max} + l^{D,\max}}{\mu_{\text{Lin}}(t_{\text{row}}, t_{\text{col}}, t_{\text{red}})} + \frac{l^{P,\max} + l^{D,\max}}{\mu_{\text{nLin}}} + N \sum_{i=1}^{l^{P,\max}+l^{D,\max}} T^{D,SA}(i) \right]. \quad (14)$$

Then, we can prove the following lower bound on the number of pending requests in the system.

Lemma 1. *Consider a system with any resource planner and scheduler. Let $\mathcal{A}^{\text{cont.}}(t)$ denote the set of requests that have arrived by time $t$. Let the "optimal request completion time" function $T_*^R(\cdot)$ be as in (13). Let $T^{\max}$ be as in (14). Then, the number of pending requests in the system at any time $t$ is lower bounded as,*

$$Q(t) \geq \frac{\left(\frac{1}{r}\sum_{j\leq A(t)} T_*^R(j) - t\right)}{T^{\max}}.$$

Proof. The proof is by contradiction. Suppose there is a time time $t$ when,

$$Q(t) < \frac{\left(\frac{1}{r}\sum_{j\leq A(t)} T_*^R(j) - t\right)}{T^{\max}}.$$

If $\frac{1}{r}\sum_{j\leq A^{\text{cont.}}(t)} T_*^R(j) \leq t$, this immediately leads to a contradiction since it implies $Q(t) < 0$. The number of pending requests in the system has to be non-negative.

Consider $\frac{1}{r}\sum_{j\leq A^{\text{cont.}}(t)} T_*^R(j) \leq t$. In this case, the worst-case amount of time taken to complete the $Q(t)$ requests happens is, a) the prompt and output lengths are the longest possible, b) none of the requests have started processing by time $t$, and c) the iterations of each request are scheduled without batching. Due to (14), the worst-case amount of time taken to complete the $Q(t)$ requests is $Q(t)T^{\max}$. This implies that the amount of time it took to complete all the requests in $\mathcal{A}^{\text{cont.}}(t)$ is,

$$\begin{aligned} T(\mathcal{A}^{\text{cont.}}(t)) &\leq t + Q(t)T^{\max}, \\ &< t + \frac{\left(\frac{1}{r}\sum_{j\leq A(t)} T_*^R(j) - t\right)}{T^{\max}} T^{\max}, \\ &< \frac{1}{r}\sum_{j\leq A(t)} T_*^R(j). \end{aligned}$$

This is a contradiction to Proposition 1 which lowers bound the time to complete a set of requests for any pair of resource planner and scheduler.

This completes the proof by contradiction. □

### B.3 Proof of Theorem 1

Recall that $\left(\mathcal{A}^{\text{cont.}}(t)\right)_{t\geq 0}$ denotes the arrival process of requests to the system, where $\mathcal{A}^{\text{cont.}}(t)$ denotes the set of requests that have arrived by time $t$. And, $A(t) \triangleq \left|\mathcal{A}^{\text{cont.}}(t)\right|$ denotes the number of arrivals by time $t$. By the premise of the Theorem, $A(t)$ is a Poisson point process of rate $\lambda$. Therefore, by the strong law of large numbers,

$$\lim_{t\to\infty} \frac{A(t)}{t} = \lambda, \quad \text{almost surely.} \tag{15}$$

Consider the sequence of requests $(R_j)_{j=1}^{\infty}$, such that $L_j^P, L_j^D \sim p_{L^P,L^D}$ i.i.d., across requests. Then, again by the strong law of large numbers,

$$\begin{aligned}\lim_{k\to\infty} \frac{1}{k}\sum_{j=1}^{k} T_*^R(j) &= \mathbb{E}\left[T_*^R(1)\right], \quad \text{almost surely,}\\ &\overset{(a)}{=} \bar{T}^R,\end{aligned}$$

where (a) follows from the formula for $T_*^R()$ in (13) and the expression for $\bar{T}^R$ in Theorem 1.

For any $t \geq 0$, requests $\left(R_j\right)_{j=1}^{A(t)}$ arrive by time $t$. Moreover, since $A(t)$ is a Poisson point process, $A(t) \longrightarrow \infty$ almost surely as $t \longrightarrow \infty$. Therefore,

$$\lim_{t\to\infty} \frac{1}{A(t)} \sum_{j\leq A(t)} T_*^R(j) = \bar{T}^R, \quad \text{almost surely.} \tag{16}$$

Combining (15) and (16), we get,

$$\begin{aligned}\lim_{t\to\infty} \frac{1}{t}\sum_{j\leq A(t)} T_*^R(j) &= \lim_{t\to\infty}\left(\frac{1}{A(t)}\sum_{j\leq A(t)} T_*^R(j)\right)\left(\frac{A(t)}{t}\right),\\ &= \left(\lim_{t\to\infty}\frac{1}{A(t)}\sum_{j\leq A(t)} T_*^R(j)\right)\left(\lim_{t\to\infty}\frac{A(t)}{t}\right),\\ &= \lambda\bar{T}^R.\end{aligned} \tag{17}$$

From Lemma 1 we have,

$$\frac{1}{t}Q(t) \geq \frac{\left(\frac{1}{r}\frac{1}{t}\sum_{j\leq A(t)} T_*^R(j) - 1\right)}{T^{\max}}.$$

Taking the limit and substituting (17), we get,

$$\begin{aligned}\lim_{t\to\infty}\frac{1}{t}Q(t) &\geq \frac{\left(\frac{1}{r}\left(\lim_{t\to\infty}\frac{1}{t}\sum_{j\leq A(t)} T_*^R(j)\right) - 1\right)}{T^{\max}},\\ &= \frac{\frac{\lambda\bar{T}^R}{r} - 1}{T^{\max}}, \quad \text{almost surely.}\end{aligned}$$

Since, $\lambda\bar{T}^R > r$, the right hand side of the above equation is positive. Therefore, substituting $\alpha = \frac{\frac{\lambda\bar{T}^R}{r}-1}{T^{\max}}$ completes the proof of Theorem 1.

## C Throughput Optimality Of RAD Proof

This section contains the proof of Theorem 2. The outline of the proof is as follows:

(1) In Section C.1, we construct a discrete-time process that tracks the number of pending requests at a node at the start of cycles of RAD. We show that this forms a discrete-time Markov chain (DTMC).
(2) In Section C.2, we bound the expected cycle times for RAD when there are at least $n$ pending requests at the start of the cycle.
(3) Building on this, in Section C.3, we show that under the conditions in Theorem 2, the DTMC tracking the number of requests at the start of cycles is positive-recurrent.
(4) Finally, in Section C.4 we show that the number of pending requests at any node forms a positive-recurrent regenerative process by showing that the expected cycle time is bounded.

### C.1 The Discrete Time Markov Chain (DTMC) Model

Here, we analyze the discrete-time process that tracks the number of pending requests at a node at the start of cycles. To do so, we introduce some stochastic processes. Consider node $r'$.

(1) Let $\left(\mathcal{A}^{\text{cont.}}_{r'}(t)\right)_{t\geq 0}$ denote the arrival process at the node. The arrival times form a Poisson point process of rate $\lambda/r$, and the $j^{\text{th}}$ request arrival to this node, $R_j$, has prompt and output lengths, $L^P_j, L^D_j \sim p_{L^P,L^D}$, independent of other requests.
(2) Let $(\mathcal{F}_{r'}(t))_{t\geq 0}$ denote the corresponding filtration with $\mathcal{F}_{r'}(t) = \sigma(\mathcal{A}^{\text{cont.}}_{r'}(t') : 0 \leq t' \leq t)$.
(3) Let $C^{-1}_{r'}[c]$ denote the start time of the $c^{\text{th}}$ cycle of the RAD scheduler, for a cycle $c \in \mathbb{N}$. Since the RAD scheduler only uses causal information, it can be easily shown that $C^{-1}_{r'}[c]$ is a stopping-time with respect to $(\mathcal{F}_{r'}(t))_{t\geq 0}$ for any $c$.
(4) Let $C_{r'}(t)$ denote the cycle that the RAD scheduler is in at time $t$, i.e., $C_{r'}(t) = \max\{c \in \mathbb{N} : C^{-1}_{r'}[c] \leq t\}$.
(5) Let $\mathcal{X}^{\text{cont.}}_{r'}(t)$ denote the set of pending requests at time $t$.
(6) Let $\mathcal{X}_{r'}[c] \triangleq \mathcal{X}^{\text{cont.}}_{r'}(C^{-1}_{r}[c])$ denote the set of pending requests at the start of cycle $c$.
(7) Let $\mathcal{A}_{r'}[c]$ denote the set of arrivals that happen during cycle $c$. That is, it is the set of arrivals in $\left(\mathcal{A}^{\text{cont.}}_{r'}(t)\right)_{t\geq 0}$ between times $\left[C^{-1}_{r'}[c], C^{-1}_{r'}[c+1]\right)$.
(8) Let $\mathcal{R}_{r'}[c]$ denote the set of requests that complete in cycle $c$.
(9) Let $Q_{r'}(t) \triangleq \left|\mathcal{X}^{\text{cont.}}_{r'}(t)\right|$ denote the number of pending requests at the node at time $t$.
(10) Let $X_{r'}[c] = |\mathcal{X}_{r'}[c]|$ denote the number of pending requests at the node at the start of cycle $c$.

We prove the following.

Lemma 2. *$(X_{r'}[c])_{c=1}^{\infty}$ is a DTMC. That is,*

$$P\left(X_{r'}[c+1] = x[c+1] \,|\, X_{r'}[c] = x[c], X_{r'}[c-1] = x[c-1], \dots\right) = P\left(X_{r'}[c+1] = x[c+1] \,|\, X_{r'}[c] = x[c]\right),$$

*for any $c \in \mathbb{N}$ and $x[0], \dots, x[c+1] \in \mathbb{N}$.*

The result will hold true due to the following three reasons,

(1) The arrivals follow a Poisson point process, making future arrivals independent of past arrivals.
(2) The RAD scheduler is "memoryless" across cycles. It does not use information about past cycles in order to schedule the requests in the latest cycle.
(3) The prompt and output lengths are i.i.d., across requests, and the RAD scheduler chooses requests in FCFS. Therefore, after knowing the number of requests at the start of the latest cycle, the prompt and output lengths of these requests are independent of the number of requests in previous cycles.

We provide a formal proof below.

Proof. Since $C_{r'}^{-1}[c]$ is a stopping time, by the strong Markov property of the Poisson point process, we have,

$$\left(\mathcal{A}_{r'}^{\text{cont.}}(t)\right)_{t>C_{r'}^{-1}[c]} \perp\!\!\!\perp \mathcal{F}_{r'}(C_{r'}^{-1}[c]), \quad \forall c \in \mathbb{N}, \tag{18}$$

where $\perp\!\!\!\perp$ denotes statistical independence. Since RAD is a causal scheduler, we have that $\mathcal{X}_{r'}^{\text{cont.}}(t)$ is causal, i.e., it is measurable in $\mathcal{F}_{r'}(t)$. This is denoted as $\mathcal{X}_{r'}^{\text{cont.}}(t) \in_m \mathcal{F}_{r'}(t)$. Therefore, we have,

$$\mathcal{X}_{r'}[c] \in_m \mathcal{F}_{r'}(C_{r'}^{-1}[c]), \quad \forall c \in \mathbb{N}, \tag{19}$$

In the $c^{\text{th}}$ cycle, the set of requests completed by the RAD scheduler is a function of the set of requests at the start of the cycle, $\mathcal{X}_{r'}[c]$, and arrival pattern after the start of the cycle, $\left(\mathcal{A}_{r'}^{\text{cont.}}(t)\right)_{t>C_{r'}^{-1}[c]}$. Therefore, the start time of the $(c+1)^{\text{th}}$ cycle, $C_{r'}^{-1}[c+1]$, the sets of arrived and completed requests in the $c^{\text{th}}$ cycle, $\mathcal{A}_{r'}[c]$ and $\mathcal{R}_{r'}[c]$, are all measurable functions of,

(1) the set of requests at the node at the start of the $c^{\text{th}}$ cycle $\mathcal{X}_{r'}[c]$,
(2) the arrival process after the start of the $c^{\text{th}}$ cycle, $\left(\mathcal{A}_{r'}^{\text{cont.}}(t)\right)_{t>C_{r'}^{-1}[c]}$.

Moreover, $\mathcal{X}_{r'}[c+1] = \mathcal{X}_{r'}[c] \bigcup \mathcal{A}_{r'}[c] \setminus \mathcal{R}_{r'}[c]$. So, for some function, $f$, that is measurable with respect to $\sigma\left(\mathcal{X}_{r'}[c], \left(\mathcal{A}_{r'}^{\text{cont.}}(t)\right)_{t>C_{r'}^{-1}[c]}\right)$,

$$\mathcal{X}_{r'}[c+1] = f\left(\mathcal{X}_{r'}[c], \left(\mathcal{A}_{r'}^{\text{cont.}}(t)\right)_{t>C_{r'}^{-1}[c]}\right). \tag{20}$$

Recall, $X_{r'}[c] = |\mathcal{X}_{r'}[c]|$. From (19), for any $c' < c$, we have $X_{r'}[c'] \in_m \mathcal{F}_{r'}\left(C_{r'}^{-1}[c]\right)$. Therefore, from (20) and the strong Markov property in (18), we can establish the following property,

$$P\left(X_{r'}[c+1] = x[c+1] \mid \mathcal{X}_{r'}[c] = \mathcal{S}, X_{r'}[c-1] = x[c-1], \dots\right) = \\ P\left(X_{r'}[c+1] = x[c+1] \mid \mathcal{X}_{r'}[c] = \mathcal{S}\right),$$

for any $c \in \mathbb{N}$ and $\boldsymbol{x}[0], \dots, \boldsymbol{x}[c+1] \in \mathbb{N}$ and any set of requests $\mathcal{S}$. Also, since $P(X = x|Y = y, g(Y) = g(y)) = P(X = x|Y = y)$ for any random variables $X$ and $Y$ and measurable function $g(\cdot)$, we can add $X_{r'}[c] = |\mathcal{S}|$ to the conditioning on both sides,

$$P\left(X_{r'}[c+1] = x[c+1] \mid \mathcal{X}_{r'}[c] = \mathcal{S}, X_{r'}[c] = |\mathcal{S}|, X_{r'}[c-1] = x[c-1], \dots\right) = \\ P\left(X_{r'}[c+1] = x[c+1] \mid \mathcal{X}_{r'}[c] = \mathcal{S}, X_{r'}[c] = |\mathcal{S}|\right), \tag{21}$$

Since RAD chooses requests FCFS, given $X_{r'}[c] = x$, $\mathcal{X}_{r'}[c]$ is simply the set of the last $x$ requests in $\left(\mathcal{A}_{r'}^{\text{cont.}}(t)\right)_{t\le C_{r'}^{-1}[c]}$. Given the cardinality of $\mathcal{X}_{r'}[c]$, which is $X_{r'}[c]$, the prompt and output lengths of the requests in $\mathcal{X}_{r'}[c]$ is i.i.d., by the premise of the Theorem. Therefore, we have,

$$P\left(\mathcal{X}_{r'}[c] = \mathcal{S} \mid X_{r'}[c] = |\mathcal{S}|\right) = \prod_{j:R_j \in \mathcal{S}} p_{L^P, L^D}\left(l_j^P, l_j^D\right) \tag{22}$$

where $l_j^P$ and $l_j^D$ denotes the prompt and output length realizations of request $R_j$.

Again, since given $X_{r'}[c] = x$, $\mathcal{X}_{r'}[c]$ is simply the set of the last $x$ requests in $\left(\mathcal{A}_{r'}^{\text{cont.}}(t)\right)_{t\le C_{r'}^{-1}[c]}$, we have that $\mathcal{X}_{r'}[c]$ is independent of the number of requests at the start of previous cycles. Therefore we have,

$$P\left(\mathcal{X}_{r'}[c] = \mathcal{S} \mid X_{r'}[c] = |\mathcal{S}|, X_{r'}[c-1] = x[c-1], \dots\right) = \prod_{j:R_j \in \mathcal{S}} p_{L^P, L^R}(l_j^P, l_j^D), \tag{23}$$

Then, multiplying (22) on the right-hand-side of (21) and multiplying (23) on the left-hand-size, and integrating over all sets $\mathcal{S}$ of size $x[c]$, we obtain the desired Markov property,

$$P\left(X_{r'}[c+1]=x[c+1] \mid X_{r'}[c]=x[c], X_{r'}[c-1]=x[c-1], \ldots\right)= \\ P\left(X_{r'}[c+1]=x[c+1] \mid X_{r'}[c]=x[c]\right),$$

□

## C.2 Cycle Times

The RAD scheduler in Algorithm 1 proceeds in cycles. We will bound the expected time of a cycle given that there are at least $n$ requests at the start of the cycle whose prompt and output lengths are distributed according to $p_{L^P,L^D}$. Recall that in this case RAD starts and completes $n$ requests chosen in FCFS manner in the cycle.

Let the set of $n$ chosen requests be denoted by $\mathcal{R}=\{R_1,\ldots,R_n\}$ (where the indexing is without loss of generality). Consider a request $R_j \in \mathcal{R}$. Recall that, due to Assumption 3, $t^*_{\text{lcm}}=\text{LCM}(t^*_{\text{row}}, t^*_{\text{col}}, t^*_{\text{red}})$ divides the prompt length $L^P_j$. In this case, the RAD scheduler schedules chunks of size $t^*_{\text{lcm}}$ for the request, $R_j$ alone in a batch. Let,

$$\mathcal{I}_j=\left\{(1,t^*_{\text{lcm}}),(t^*_{\text{lcm}}+1,t^*_{\text{lcm}}),\ldots,(L^P_j-t^*_{\text{lcm}},t^*_{\text{lcm}})\right\}$$

denote the set of chunk positions and chunk sizes scheduled to complete the prefill-phase. Then, required time to complete its prefill-phase, once scheduled, is,

$$\begin{aligned} T^P_j &= \frac{1}{\mu_{\text{Lin}}(t^*_{\text{row}}, t^*_{\text{col}}, t^*_{\text{red}})}\frac{L^P_j}{t^*_{\text{col}}}+\frac{1}{\mu_{\text{nLin}}}L^P_j \\ &+\frac{N}{s\mu(t^*_{\text{row}}, t^*_{\text{col}}, t^*_{\text{red}})}\sum_{(i_j,c_j)\in\mathcal{I}_j}\left(\frac{i_j+c_j-1}{t^*_{\text{row}}}\frac{c_j}{t^*_{\text{col}}}\frac{d}{t^*_{\text{red}}}+\frac{d}{t^*_{\text{row}}}\frac{c_j}{t^*_{\text{col}}}\frac{i_j+c_j-1}{t^*_{\text{red}}}\right), \end{aligned}$$

Taking the sum and the expectation, the expected amount of time to complete the prefill-phases of all the $n$ requests can be expressed as,

$$\begin{aligned} \mathbb{E}\left[T^{PM}(\mathcal{R})\right] = n\cdot\Bigg[&\frac{1}{\mu_{\text{Lin}}(t^*_{\text{row}}, t^*_{\text{col}}, t^*_{\text{red}})}\frac{\mathbb{E}\left[L^P\right]}{t^*_{\text{col}}}+\frac{\mathbb{E}\left[L^P\right]}{\mu_{\text{nLin}}} \\ &+\frac{Nd}{s\mu(t^*_{\text{row}}, t^*_{\text{col}}, t^*_{\text{red}})t^*_{\text{row}}t^*_{\text{col}}t^*_{\text{red}}}\mathbb{E}\left[L^P(L^P+t^*_{\text{lcm}})\right]\Bigg]. \end{aligned} \tag{24}$$

Next, we will focus on bounding the effective execution time of decode-iterations. For this, consider the set of batch indices, $\mathcal{M}^D$, where only decode-iterations are scheduled, and denote the batches by $\left(\mathcal{B}^D_m\right)_{m\in\mathcal{M}^D}$. Moreover, let $\mathcal{M}^D_*$ denote the subset of these batch indices that contain exactly $t^*_{\text{col}}$ decode-iterations. In other words, these batches are optimally tiled. Therefore, for $m\in\mathcal{M}^D_*$ we have the batch execution time given as,

$$T^B_*(\mathcal{B}^D_m)=\frac{1}{\mu_{\text{Lin}}(t^*_{\text{row}}, t^*_{\text{col}}, t^*_{\text{red}})}\frac{t^*_{\text{col}}}{t^*_{\text{col}}}+\frac{1}{\mu_{\text{nLin}}}t^*_{\text{col}}+N\sum_{j:R_j\in\mathcal{S}^D(\mathcal{B}^D_m)}T^{D,SA}(i_j).$$

where recall that $i_j$ is the token position corresponding to the decode-iteration of request $R_j$ in the batch.

For all other batches $m \in \mathcal{M}^D \setminus \mathcal{M}^D_*$, the batch execution times are upper bounded as,

$$T^B(\mathcal{B}^D_m) \leq \max_{(t_{\text{row}}, t_{\text{col}}) \in \mathcal{T}_{\text{out}}, t_{\text{red}} \in \mathcal{T}_{\text{out}}} \frac{1}{\mu_{\text{Lin}}(t_{\text{row}}, t_{\text{col}}, t_{\text{red}})} \left\lceil \frac{|\mathcal{B}^D_m|}{t_{\text{col}}} \right\rceil + \frac{1}{\mu_{\text{nLin}}} |\mathcal{B}^D_m| + N \sum_{j: R_j \in \mathcal{S}^D(\mathcal{B}^D_m)} T^{D,SA}(i_j).$$

Let $\mathcal{R}^*$ denote the set of requests whose every decode iteration is scheduled in an optimally tiled batch. For $R_{j^*} \in \mathcal{R}^*$, denote the "effective completion time" it took to complete its decode-phase as,

$$T^D_*(j^*) = \frac{1}{\mu_{\text{Lin}}(t^*_{\text{row}}, t^*_{\text{col}}, t^*_{\text{red}})} \frac{L^D_{j^*}}{t^*_{\text{col}}} + \frac{1}{\mu_{\text{nLin}}} L^D_{j^*} + N \sum_{i=L^P_{j^*}+1}^{L^P_{j^*}+L^D_{j^*}} T^{D,SA}(i).$$

For all other requests, $R_{j'} \in \mathcal{R} \setminus \mathcal{R}^*$, their "effective completion times" can be bounded as,

$$T^{D,\max}(j') \leq \max_{(t_{\text{row}}, t_{\text{col}}) \in \mathcal{T}_{\text{out}}, t_{\text{red}} \in \mathcal{T}_{\text{out}}} \frac{1}{\mu_{\text{Lin}}(t_{\text{row}}, t_{\text{col}}, t_{\text{red}})} L^D_{j'} + \frac{1}{\mu_{\text{nLin}}} L^D_{j'} + N \sum_{i=L^P_{j'}+1}^{L^P_{j'}+L^D_{j'}} T^{D,SA}(i).$$

Due to Assumption 3,

$$T^{D,\max}(j') \leq T^{\max}, \quad \text{almost surely,}$$

where $T^{\max}$ is as defined in Theorem 2.

The condition in Line 6 of Algorithm 1 ensures that when $n$ requests are being processed in the cycle, i.e., $|\mathcal{R}| = n$, then all the decode-iterations of at least $n - t^*_{\text{col}} + 1$ requests are optimally tiled, i.e., $|\mathcal{R}^*| \geq n - t^*_{\text{col}}$. Therefore, $|\mathcal{R} \setminus \mathcal{R}^*| \leq t^*_{\text{col}} - 1$.

The amount of time required to complete the decode-phase of requests in this cycle can thus be bounded as,

$$\begin{aligned}
T^{DM} &= \sum_{m \in \mathcal{M}^D} T^B(\mathcal{B}^D_m), \\
&= \sum_{m \in \mathcal{M}^D_*} T^B(\mathcal{B}^D_m) + \sum_{m \in \mathcal{M}^D \setminus \mathcal{M}^D_*} T^B(\mathcal{B}^D_m), \\
&\leq \sum_{j^*: R_{j^*} \in \mathcal{R}^*} T^D_*(j^*) + \sum_{j': R_{j'} \in \mathcal{R} \setminus \mathcal{R}^*} T^{D,\max}(j'), \\
&\leq \sum_{j: R_j \in \mathcal{R}} T^D_*(j) + \sum_{j: R_{j'} \in \mathcal{R} \setminus \mathcal{R}^*} T^{D,\max}(j'), \\
&\leq \sum_{j: R_j \in \mathcal{R}} T^D_*(j) + (t^*_{\text{col}} - 1) T^{\max}.
\end{aligned}$$

Since each request has prompt and output length distributed according to $p_{L^P, L^D}$, the expected time to complete the decode-phase of all the requests in the cycle can be upper bounded as,

$$\begin{aligned}
\mathbb{E}\left[T^{DM}\right] \leq n &\left( \frac{1}{\mu_{\text{Lin}}(t^*_{\text{row}}, t^*_{\text{col}}, t^*_{\text{red}})} \frac{\mathbb{E}\left[L^D\right]}{t^*_{\text{col}}} + \frac{1}{\mu_{\text{nLin}}} \mathbb{E}\left[L^D\right] + N \mathbb{E}\left[ \sum_{i=L^P+1}^{L^P+L^D} T^{D,SA}(i) \right] \right) \\
&+ (t^*_{\text{col}} - 1) T^{\max}.
\end{aligned} \tag{25}$$

Therefore, the expected cycle time is upper bounded as,

$$\begin{aligned}
\mathbb{E}\left[T^{\text{cycle}}\right] &= \mathbb{E}\left[T^{PM}\right] + \mathbb{E}\left[T^{DM}\right], \\
&\leq n \bar{T}^R + (t^*_{\text{col}} - 1) T^{\max},
\end{aligned} \tag{26}$$

where the upper bound is obtained by combining (24), (25), and the definition of $\bar{T}^R$ from Theorem 1.

### C.3 Positive recurrence of the DTMC

Lemma 3. *Under the same conditions as in Theorem 2,* $(X_{r'}[c])_{c=1}^{\infty}$ *is positive recurrent.*

Proof. Consider a state $X_{r'}[c]$ with $X_{r'}[c] \geq n$. Denote $A_{r'}[c] = |\mathcal{A}_{r'}[c]|$ to be the number of request arrivals that happen during the $c^{\text{th}}$ cycle of RAD. Then, $X_{r'}[c+1] = X_{r'}[c] - n + A_{r'}[c]$, since RAD completes exactly $n$ requests in a cycle if there are at least $n$ requests at the start of the cycle.

Since there are already at least $n$ requests at the start of the cycle, and RAD chooses requests in a FCFS manner, the set of requests completed in the cycle is a subset of these requests. That is,

$$\mathcal{R}_{r'}[c] \subset \mathcal{X}_{r'}[c] \quad \text{and} \quad |R_{r'}[c]| = n, \quad \text{if } X_{r'}[c] \geq n.$$

Therefore, in this case, the duration of this cycle is independent of the arrivals after the cyle starts, i.e.,

$$C_{r'}^{-1}[c+1] - C_{r'}^{-1}[c] \perp\!\!\!\perp \left(\mathcal{A}_{r'}^{\text{cont.}}(t)\right)_{t > C_{r'}^{-1}[c]}, \quad \text{if } X_{r'}[c] \geq n.$$

Requests arrive according to a Poisson point process of rate $\lambda/r$. Therefore, by the independence of the cycle duration to the arrival process, we have the expected number of arrivals during the cycle given as,

$$\mathbb{E}\left[A_{r'}[c]\right] = \lambda \mathbb{E}\left[C_{r'}^{-1}[c+1] - C_{r'}^{-1}[c]\right], \quad \text{if } X_{r'}[c] \geq n.$$

From the upper bound on the expected duration of the cycle in (26), we have,

$$\mathbb{E}\left[A_{r'}[c]\right] \leq \frac{\lambda}{r}\left(n\bar{T}^R + (t_{\text{col}}^* - 1)T^{\max}\right), \quad \text{if } X_{r'}[c] \geq n.$$

Then the drift for all the states $x$ with $x \geq n$ is,

$$\begin{aligned}
\mathbb{E}\left[X_{r'}[c+1] - X_{r'}[c] \middle| X_{r'}[c] = x\right] &= n\left(\frac{\lambda}{r}\bar{T}^R - 1\right) + \lambda(t_{\text{col}}^* - 1)T^{\max}, \\
&\overset{(a)}{\leq} -n\varepsilon + (t_{\text{col}}^* - 1)\lambda T^{\max}, \\
&\overset{(b)}{<} 0, \quad \forall c \in \mathbb{N}.
\end{aligned}$$

(a) follows by the premise, $\lambda\bar{T}^R \leq r(1-\varepsilon)$, and (b) follows from the choice of $n$ in Theorem 2.

And clearly $\mathbb{E}\left[X_{r'}[c+1] \middle| X_{r'}[c] = x\right] < \infty$ for any $x \in \mathbb{N}$.

Therefore, by Foster's Theorem[6], $(X_{r'}[c])_{c=1}^{\infty}$ is positive recurrent. □

### C.4 Proof of Theorem 2

First we show that the expected cycle times are finite.

Lemma 4. *Under the same conditions as Theorem 2, for any cycle index* $c \in \mathbb{N}$*:*

$$\mathbb{E}\left[C_{r'}^{-1}[c+1] - C_{r'}^{-1}[c]\right] \leq \frac{r}{\lambda} + nT^{\max}.$$

Proof. Consider $X_{r'}[c] > 0$. Then, the RAD scheduler completes at most $n$ requests in this cycle. The worst case time that a scheduler may take to complete $n$ requests is when, a) the prompt and output lengths of all requests are the longest possible, and, b) the scheduler processes the requests one after the other. $T^{\max}$ bounds the worst-case time required to complete a request. Therefore, the worst-case cycle time of the RAD scheduler is bounded by $nT^{\max}$.

Consider $X_{r'}[c] = 0$. In this case, the RAD scheduler waits for a request to arrive, which takes expected time $r/\lambda$. Then, similar to above, the RAD scheduler may complete up to $n$ requests (in the case where more requests arrive before RAD proceeds to the next cycle). We know that that worst-case cycle time to complete $n$ requests is $nT^{\max}$. Therefore, in this case, the expected cycle time is bounded as $r/\lambda + nT^{\max}$. □

Finally, we using the previous Lemmas proved in this section, we present the proof of Theorem 2.

Proof of Theorem 2. By definition, $Q_{r'}\left(C^{-1}[c]\right) = 0$, if and only if $X_{r'}[c] = 0$. And, by Lemma 3, $(X_{r'}[c])_{c=1}^{\infty}$ is positive-recurrent. And, by Lemma 4, for any $c_1, c_2 \in \mathbb{N}$ such that $c_1 < c_2$,

$$\mathbb{E}\left[C_{r'}^{-1}[c_2] - C_{r'}^{-1}[c_1]\right] \leq \frac{(c_2 - c_1)r}{\lambda} + (c_2 - c_1)nT^{\max}.$$

We know the systems starts empty: $Q_{r'}(0) = 0$. Define, $T^{\text{regen}} = \min\{t > 0 : Q_{r'}(t) = 0\}$. Then by the above results, $\mathbb{E}[T^{\text{regen}}] < \infty$. Moreover, RAD starts a new cycle whenever $Q_{r'}(t) = 0$. Therefore, $T^{\text{regen}}$ is aligned with the start of a new cycle: $T^{\text{regen}} = C^{-1}[C(T^{\text{regen}})]$. Since $(X_{r'}[c])_{c=0}^{\infty}$ is a DTMC, this implies that $(Q_{r'}(t))_{t>T^{\text{regen}}}$ is independent of $(Q_{r'}(t))_{t \leq T^{\text{regen}}}$. Therefore, $(Q_{r'}(t))_{t \geq 0}$ is a regenerative process, with regeneration point 0. Further it is positive recurrent since $\mathbb{E}[T^{\text{regen}}] < \infty$. □

## D Another Throughput-Optimal Scheduler

Another throughput-optimal scheduler is presented in Algorithm 2. It has the same high level design principle as the RAD scheduler. It proceeds in cycles, where in each cycle it starts and completes up to $n$ requests. This scheduler differs from RAD in the order in which it schedules prefill and decode iterations.

Algorithm 2 first completes up to $n$ prefill-phase requests (if enough are available) with optimal tiling (see Lines 3-13). Then it moves to the decode mode, where it first chooses the first $t^*_{\text{col}}$ of the decode phase requests, $\mathcal{D}$, and marks them as active, denoted $\mathcal{D}_{\text{active}}$. It then proceeds to schedules batches of decode-iterations of each request in $\mathcal{D}_{\text{active}}$. After each request it removes requests that sampled their `stop` token, and replaces them with remaining requests in $\mathcal{D}$, if there are any. This proceeds until all the requests in this cycle have completed.

It may be shown that Algorithm 2 schedules optimally tiled batches most of the time, and also performs dynamic optimal resource allocation between prefill and decode phase requests just like the RAD scheduler. As such, Theorem 2 applies when the RAD scheduler is replaced by the scheduler in Algorithm 2 as well. The proof would be very similar to the one presented in Appendix C for the RAD scheduler. We omit the proof to avoid repetition.

The primary difference in the operation of the two schedulers would be the TTFT and TBT observed by requests. For the scheduler in Algorithm 2, consider that the first output token of a request is only produced when it has been moved to the active decode set, $\mathcal{D}_{\text{active}}$. In this case, the decode-iterations of this request keep getting scheduled in consecutive batches until the request is complete. Therefore, the TBT of requests will be low. However, since the scheduler has to complete a large number, $n$, of requests before it starts generating the tokens for any of the requests, the TTFT of the requests will be very high. This is complementary to the RAD scheduler, where requests observe a hight TBT, but a relatively lower TTFT.

## E Other Schedulers in the Framework

As a way of illustration, we provide a description of request-level batching, Sarathi-serve and DistServe in our scheduler framework introduced in Section 3.1.

**Algorithm 2** Another Throughput Optimal Scheduler

**Require:** Parameter $n$

$\mathcal{P} \leftarrow \emptyset, \mathcal{D} \leftarrow \emptyset$
**while** True **do**
  // Cycle starts in PREFILL MODE
  $requests_in_cycle \leftarrow 0$
  **while** $\mathcal{P} \neq \emptyset$ **and** $requests_in_cycle < n$ **do**
    Choose a request $R$ first-come-first-serve from $\mathcal{P}$ and set $i \leftarrow 1$
    **while** prefill phase of the chosen request is incomplete **do**
      $c = \min\{\text{LCM}(t^*_{\text{row}}, t^*_{\text{col}}, t^*_{\text{red}}), L_P(R) - i\}$
      Schedule $PI(R, i, c)$ in a batch.
      $i \leftarrow i + c$
    **end while**
    Move the request from $\mathcal{P}$ to $\mathcal{D}$
    $requests_in_cycle \leftarrow requests_in_cycle + 1$
  **end while**
  //Change mode to DECODE MODE
  $\mathcal{D}_{\text{active}} = \emptyset$
  Move $\min(t^*_{\text{col}}, |\mathcal{D}|)$ requests from $\mathcal{D}$ to $\mathcal{D}_{\text{active}}$
  **while** $|\mathcal{D}_{\text{active}}| > 0$ **do**
    Schedule one DI for each request in $\mathcal{D}_{\text{active}}$ in a batch.
    Remove requests that sampled the "stop" token from $\mathcal{D}_{\text{active}}$
    Move $\min(t^*_{\text{col}} - |\mathcal{D}_{\text{active}}|, |\mathcal{D}|)$ requests from $\mathcal{D}$ to $\mathcal{D}_{\text{active}}$.
  **end while**
**end while**

## E.1 Request level batching

This scheduler is inspired by FasterTransformer[23], which schedules a batch of $b$ requests at a time, and only proceeds to other requests after all these requests have completed. We adapt this scheduler and express it in the "sequences of batches of iterations" framework in Algorithm 3. Here the scheduler alternates between two modes. It starts in Prefill Mode in which it selects $b$ requests from the prefill-queue in FCFS order and completes their prefill phases. It then switches to Decode Mode where it schedules one decode iteration for all active decode-phase requests until all of them complete. It then switches back to Prefill Mode.

## E.2 Sarathi-serve

This scheduler, as shown in Algorithm 4, enforces a fixed token budget per batch and prioritizes decode-phase requests. It firsts selects a decode iteration for each decode-phase request (the number of these is guaranteed to be below the budget). It then fills the remaining capacity with chunked prefill iterations. Completed prefills are transitioned to the decode queue, while completed decodes are removed.

## E.3 DistServe

This is a distributed scheduler, as shown in Algorithm 5, and runs on dedicated prefill and decode nodes. In prefill nodes, prefill requests are handled one at a time in FCFS order, with entire prompts processed in a single iteration. Completed requests are transferred to the decode node. At the decode node, it continuously batches and schedules decode iterations. Here we note that it is

**Algorithm 3** Request-Level Batching Scheduler [23]

**Require:** Batch size $b$

```
𝒫 ← ∅, 𝒟 ← ∅                                   // Prefill and decode phase queues
mode ← Decode
while True do
    Update 𝒫 based on arrivals
    if mode = Prefill then
        if 𝒟 ≠ ∅ then
            Schedule DI for each R ∈ 𝒟 in a batch with appropriate token positions
            Remove from 𝒟 any request that sampled the stop token
        else
            mode ← Prefill
        end if
    else if mode = Prefill then
        if 𝒫 ≠ ∅ then
            Choose up to b requests from 𝒫 in FCFS order
            for all selected requests R_j do
                Schedule PI(R, 1, L_j^P)                // Full prompt in one iteration
                Move R from 𝒫 to 𝒟
            end for
        end if
        mode ← Decode
    end if
end while
```

possible to run both prefill and decode requests with optimal tiling at the prefill and decode nodes respectively.

## F Additional experimental results

### F.1 Results for the scenario of 5% split

Figure 9 presents a comparative evaluation of scheduling policies under heterogeneous TTFT and TBT constraints, with a workload comprising 5% paying users. Figure 9a shows the median TTFT for all requests as a function of request rate, plotted on a log-scaled y-axis to highlight differences at low load. This view reveals how various schedulers handle contention-free versus saturated conditions. Figure 9b reports the number of requests completed at the peak load of 1.6 requests per second, bucketed by prompt length. Notably, SLAI (SPF, dynamic offset) serves nearly as many requests as Sarathi-serve while achieving substantially lower median TTFT, whereas vLLM exhibits instability and fails to maintain throughput under high load. Finally, Figure 9c plots the mean TTFT at 1.6 requests per second as a function of prompt length. Despite favoring shorter prompts, SPF-based schedulers yield a lower overall TTFT compared to FCFS variants, demonstrating the benefit of prioritizing short requests even in the presence of heterogeneous job sizes.

### F.2 Prioritizing prefill-phase requests of paying users over free-tier users

In this section, we evaluate an additional policy: SLAI (SPF with priority, dynamic offset). This policy gives strict priority to prefill-phase requests from paying users over those from free-tier users, regardless of prompt length. In other words, it always schedules a paying user's request

**Algorithm 4** Sarathi-Serve Scheduler[2]

**Require:** Token budget $\tau^{\text{budget}}$ `// Max tokens per batch`

$\mathcal{P} \leftarrow \emptyset, \mathcal{D} \leftarrow \emptyset$ `// Requests in prefill and decode phases`
**while** True **do**
  Update $\mathcal{P}$ based on arrivals
  $\mathcal{B} \leftarrow \emptyset$ `// Current batch and used tokens`
  **for all** requests $R \in \mathcal{D}$ **do**
    Add $DI(R, i)$ to batch $\mathcal{B}$
    Mark $i \leftarrow i + 1$ for $R$
  **end for**
  $\tau \leftarrow |\mathcal{D}|$
  **for all** $R \in \mathcal{P}$ **in arrival order do**
    **while** $\tau < \tau^{\text{budget}}$ **do**
      $c \leftarrow \min\{\tau^{\text{budget}} - \tau, L_j^P - i\}$ `// i is starting token index left to prefill`
      Add $PI(R, i, c)$ to batch $\mathcal{B}$
      $\tau \leftarrow \tau + c$
      Mark $i \leftarrow i + c$ for $R$
    **end while**
  **end for**
  Schedule batch $\mathcal{B}$
  **for all** requests $R \in \mathcal{P}$ **do**
    **if** prefill of $R$ complete **then**
      Move $R$ from $\mathcal{P}$ to $\mathcal{D}$
    **end if**
  **end for**
  **for all** requests $R \in \mathcal{D}$ **do**
    **if** decode of $R$ sampled `stop` token **then**
      Remove $R$ from $\mathcal{D}$
    **end if**
  **end for**
**end while**

first. All other parameters are the same as in SLAI (SPF, dynamic offset). Figure 10 compares this priority-based policy with other scheduling strategies. At high load (1.6 requests per second), we observe that the mean TTFT for paying users is lower than that for free-tier users. However, the improvement in TTFT for paying users is relatively small.

### F.3 Results for the scenario of 50% and 95% split

In this section, we present additional results for scenarios with less heterogeneity in user workloads. Figures 11 and 12 show results similar to those discussed earlier, but for cases where the percentage of paying users is 50% and 95%, respectively. As the proportion of paying users increases, the improvement in serving capacity under SLAI (SPF, dynamic offset) compared to Sarathi-serve (FCFS) becomes smaller. This is because a larger share of traffic now has stricter TBT constraints, leaving fewer opportunities for SLAI to defer decode-phase requests dynamically.

**Algorithm 5** DistServe Scheduler [36]

**Require:** Separate compute nodes for prefill and decode phases

**— Prefill Node —**

$\mathcal{P} \leftarrow \emptyset$ `// Requests awaiting prefill`
**while** True **do**
  Update $\mathcal{P}$ based on new arrivals
  **if** $\mathcal{P} = \emptyset$ **then**
    **continue**
  **end if**
  Choose request $R_j$ from $\mathcal{P}$ in first-come-first-serve order
  Schedule $PI(R_j, 1, L_j^P)$ `// chunked prefill-iteration may be done too`
  Transfer KV-cache of $R$ to a Decode Node
  Remove $R$ from $\mathcal{P}$
**end while**

**— Decode Node —**

$\mathcal{D} \leftarrow \emptyset$ `// Requests in decode phase`
**while** True **do**
  Update $\mathcal{D}$ received from a Prefill Node
  **if** $\mathcal{D} = \emptyset$ **then**
    **continue**
  **end if**
  Schedule $DI(R)$ for each $R \in \mathcal{D}$ in a batch `//` $t^*_{\text{col}}$ `requests may also be scheduled for optimal tiling`
  Remove from $\mathcal{D}$ any request that sampled the `stop` token
**end while**

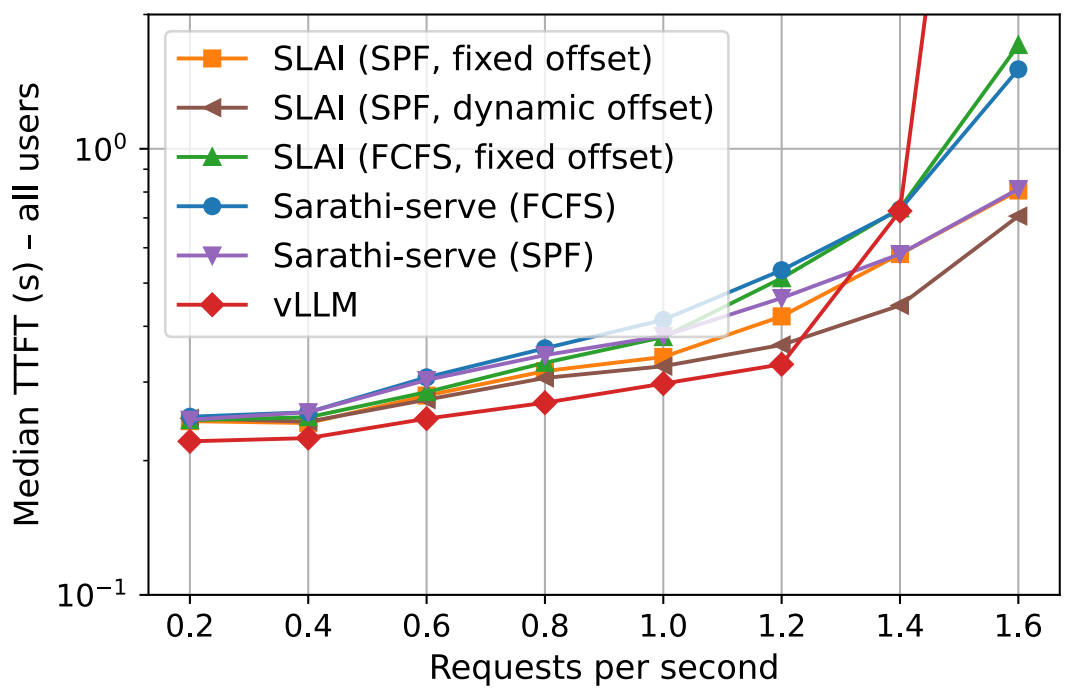

(a) Median TTFT for all users as a function of request rate, shown on a log-scaled $y$-axis to illustrate the gap at lower request rates.

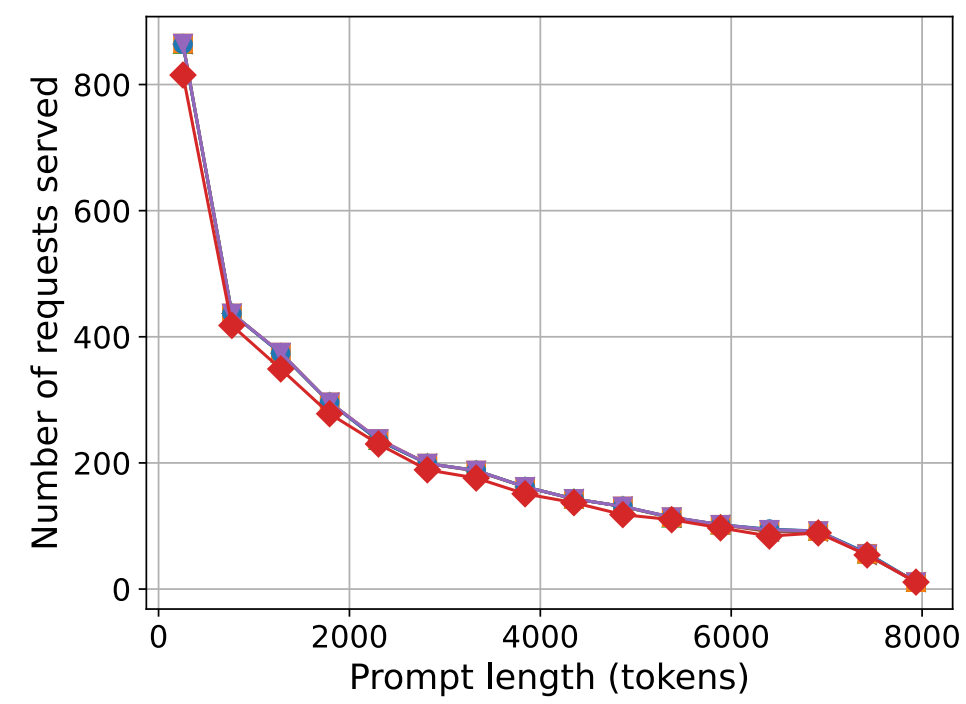

(b) Requests served at high load (1.6 req/s) versus prompt length.

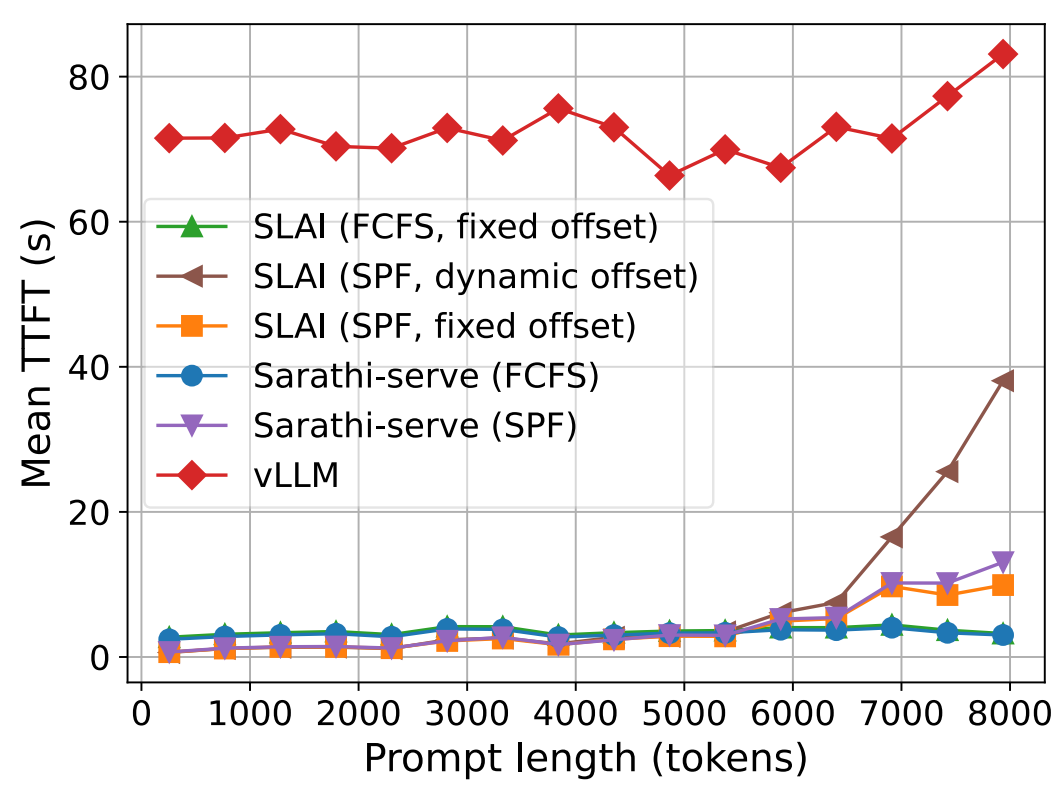

(c) Mean TTFT at the high load (1.6 req/s) versus prompt length.

Fig. 9. Performance comparison of different policies under mixed user workloads with 5% paying users.

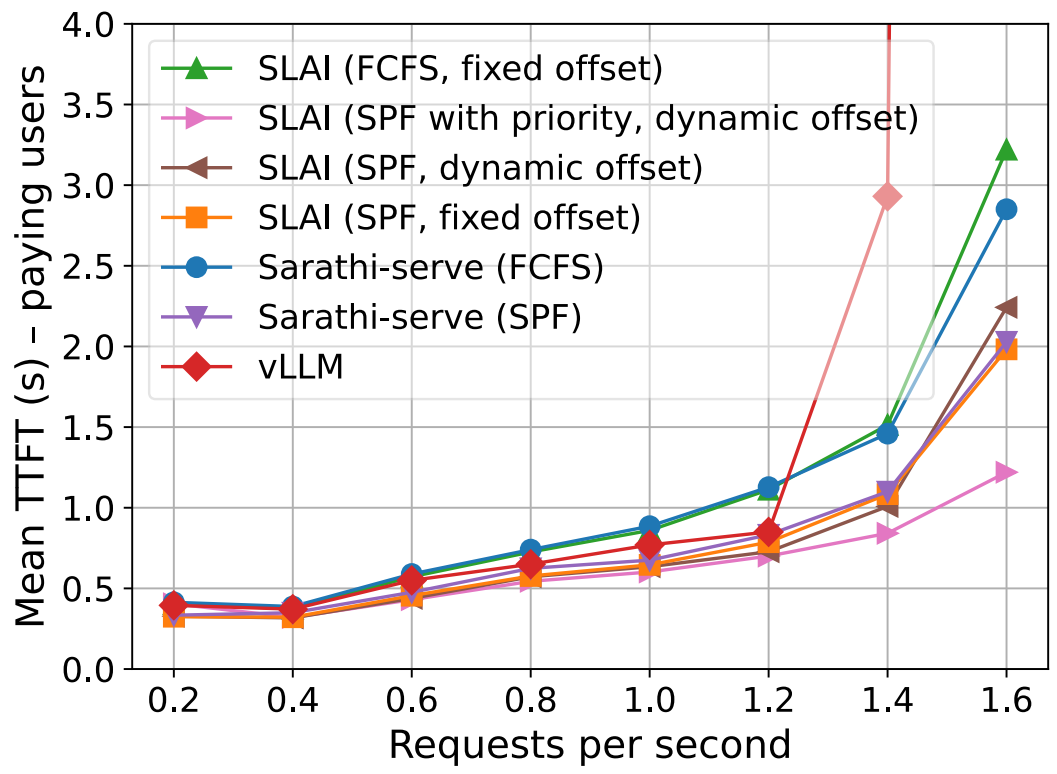

(a) Mean TTFT for paying users as a function of requests per second.

(b) Mean TTFT for free-tier users as a function of requests per second.

Fig. 10. Performance comparison of different policies under mixed user workloads with 5% paying users.

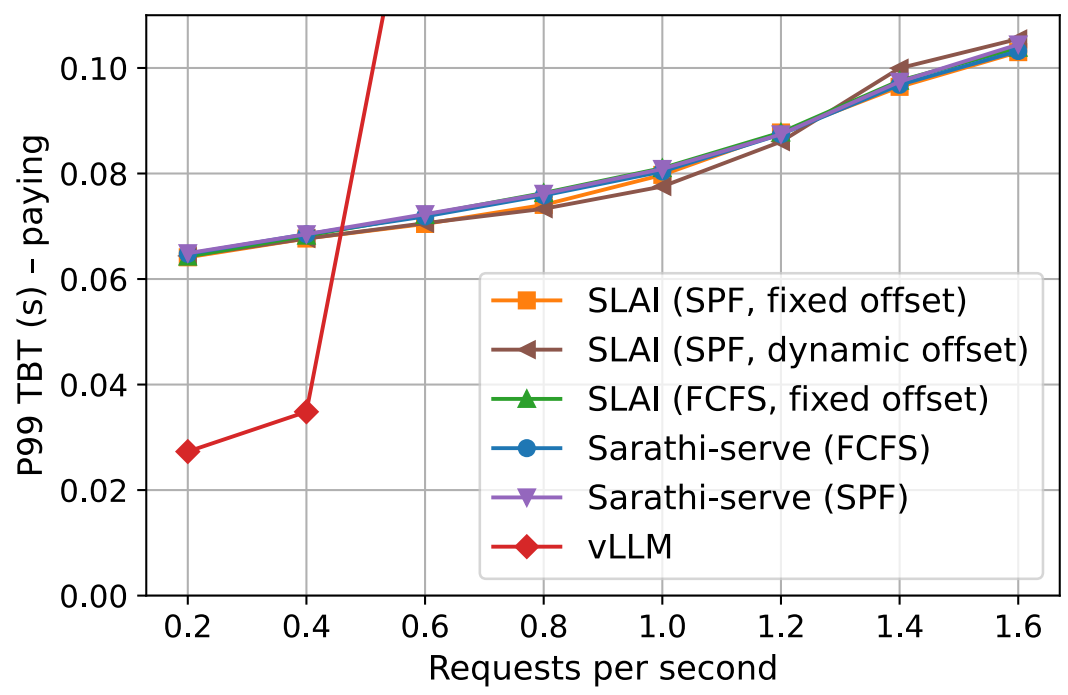

(a) 99th percentile TBT for paying users across different request rates for a target of 0.1 seconds.

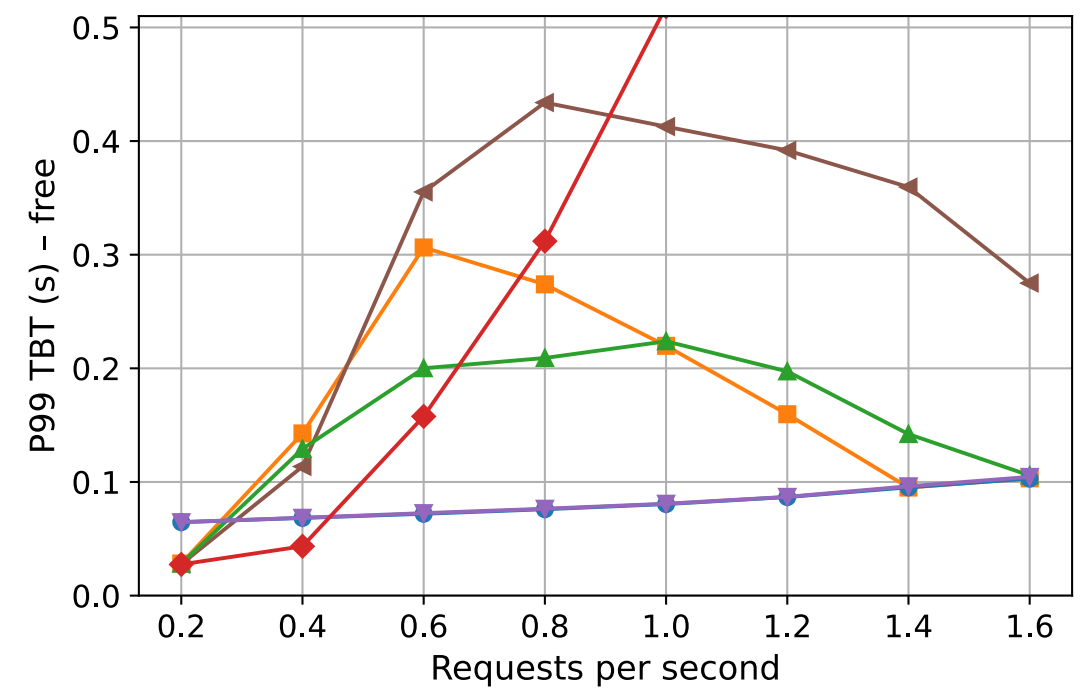

(b) 99th percentile TBT for free-tier users across different request rates for a target of 0.5 seconds.

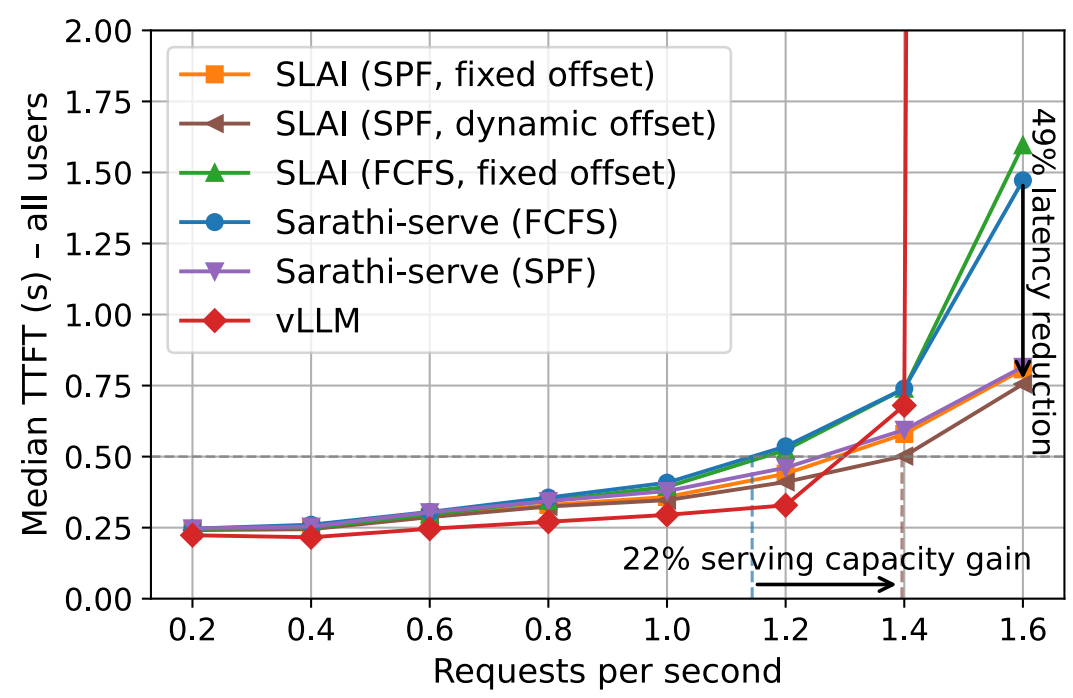

(c) Median TTFT for all users as a function of request rate. SLAI (SPF, dynamic offset) reduces TTFT from 1.5 seconds (under Sarathi-serve (FCFS)) to 0.73 seconds, and increases peak serving capacity from 1.15 to 1.4 requests per second subject to latency constraints.

Fig. 11. Performance comparison of SLAI, Sarathi-serve, and vLLM under mixed user workloads with 50% paying users. SLAI (SPF, dynamic offset) achieves the best latency-throughput trade-off.

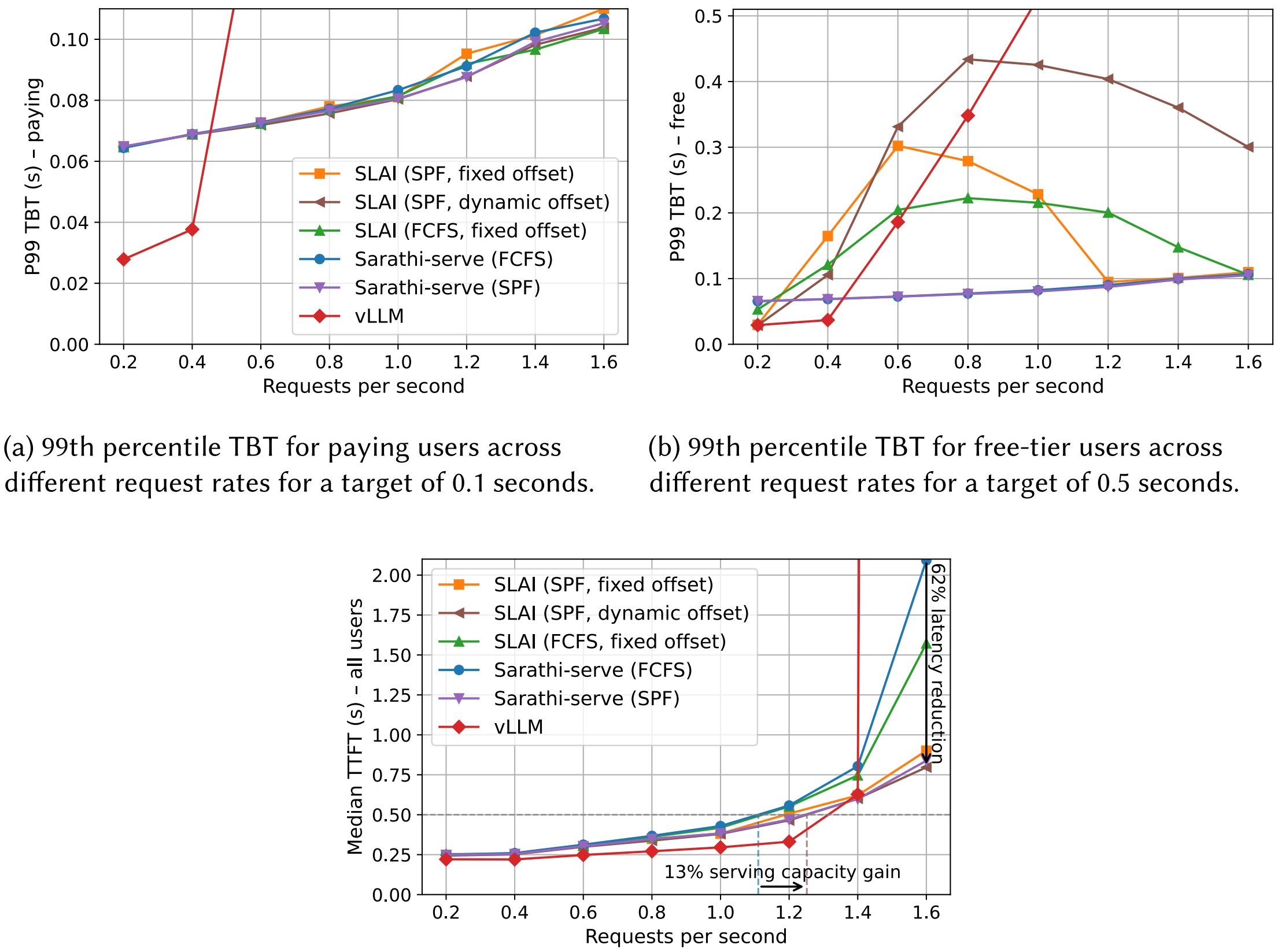

(a) 99th percentile TBT for paying users across different request rates for a target of 0.1 seconds.

(b) 99th percentile TBT for free-tier users across different request rates for a target of 0.5 seconds.

(c) Median TTFT for all users as a function of request rate. SLAI (SPF, dynamic offset) reduces TTFT from 2 seconds (under Sarathi-serve (FCFS)) to 0.75 seconds, and increases peak serving capacity from 1.15 to 1.25 requests per second subject to latency constraints.

Fig. 12. Performance comparison of SLAI, Sarathi-serve, and vLLM under mixed user workloads with 95% paying users. SLAI (SPF, dynamic offset) achieves the best latency-throughput trade-off.